%% file: main.tex
\documentclass[10pt,twocolumn,letterpaper]{article}

\usepackage{cvpr}              %

\usepackage{caption}

\usepackage{lipsum} %
\usepackage{fancyhdr}
\usepackage{xcolor} %
\usepackage{adjustbox}
\usepackage{graphicx}
\usepackage{amsmath}
\usepackage{amssymb}
\usepackage{booktabs}
\usepackage{array}
\usepackage{tabularx} 
\usepackage{afterpage}

\include{include/mymacros}

\definecolor{cvprblue}{rgb}{0.21,0.49,0.74}
\usepackage[pagebackref,breaklinks,colorlinks,citecolor=cvprblue]{hyperref}
\usepackage[capitalize]{cleveref}
\crefname{section}{Sec.}{Secs.}
\Crefname{section}{Section}{Sections}
\Crefname{table}{Table}{Tables}
\crefname{table}{Tab.}{Tabs.}

\usepackage{tocloft}

\def\paperID{312} %
\def\confName{3DV\xspace}
\def\confYear{2025\xspace}

\title{RealmDreamer: Text-Driven 3D Scene Generation with \\ Inpainting and Depth Diffusion} 

\author{
Jaidev Shriram$^1$\footnotemark[1] \and 
Alex Trevithick$^1$\footnotemark[1] \and 
Lingjie Liu$^2$ \and 
Ravi Ramamoorthi$^1$\\
\vspace{-0.4cm}
\and $^1$University of California, San Diego
\and $^2$University of Pennsylvania\\
}

\begin{document}

\renewcommand{\baselinestretch}{0.95}

\twocolumn[{%
\renewcommand\twocolumn[1][]{#1}
\vspace{-1cm}
\maketitle
\vspace{-0.7cm}
\begin{center}
    \vspace{-0.35cm}
    \large\url{https://realmdreamer.github.io/}
\end{center}
\vspace{0.01cm}

\begin{center}
    \centering
    \captionsetup{type=figure}
        \includegraphics[width=\linewidth]{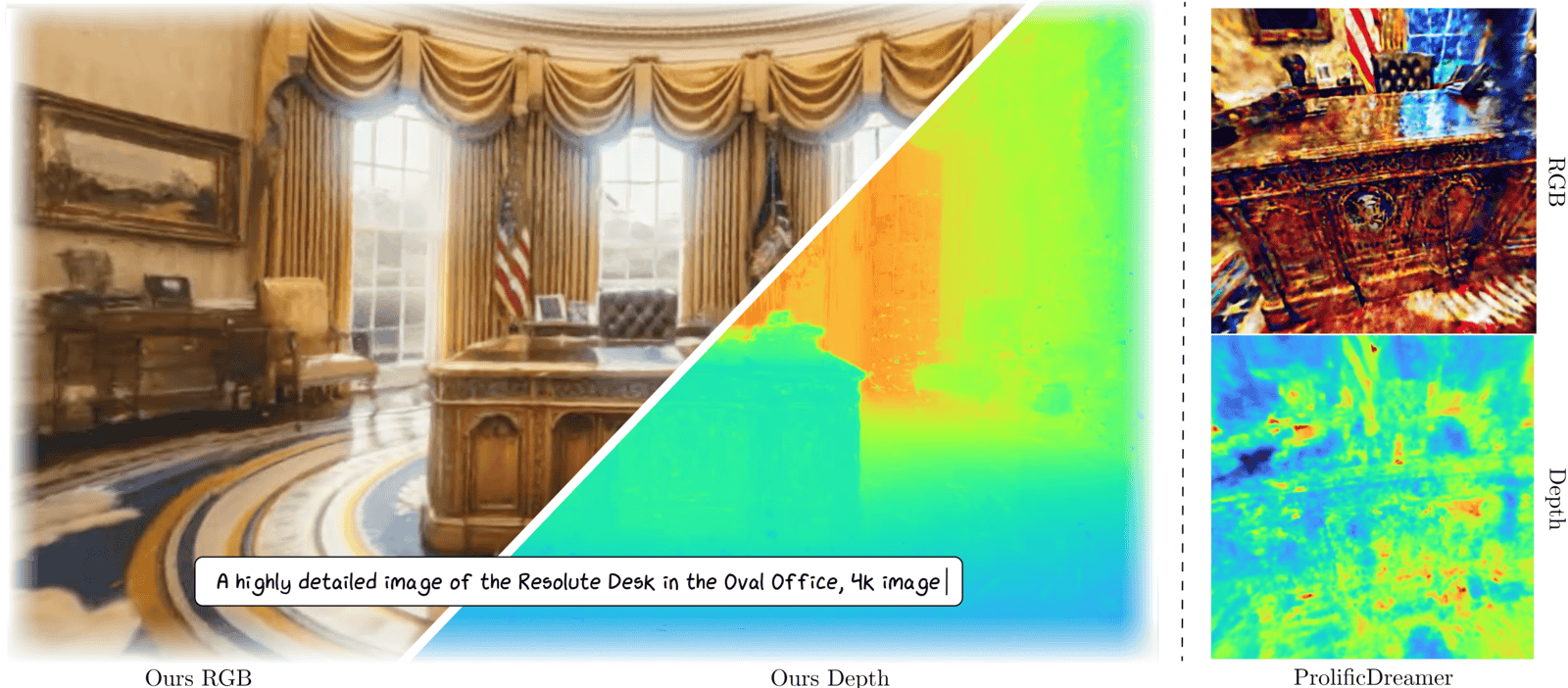}
    \captionof{figure}{\textbf{A scene created by our method on the left compared to baseline ProlificDreamer~\cite{wang2023prolificdreamer} on the right.}  \pname\ generates 3D scenes from text prompts (as above), achieving state-of-the-art results with parallax, detailed appearance, and realistic geometry.}
    \label{fig:teaser}
\end{center}%
}]

\begin{abstract}
\vspace{-0.2cm}
We introduce \textbf{\pname}, a technique for generating forward-facing 3D scenes from text descriptions. Our method optimizes a 3D Gaussian Splatting representation to match complex text prompts using pretrained diffusion models. Our key insight is to leverage 2D inpainting diffusion models conditioned on an initial scene estimate to provide low variance supervision for unknown regions during 3D distillation. In conjunction, we imbue high-fidelity geometry with geometric distillation from a depth diffusion model, conditioned on samples from the inpainting model. We find that the initialization of the optimization is crucial, and provide a principled methodology for doing so. Notably, our technique doesn't require video or multi-view data and can synthesize various high-quality 3D scenes in different styles with complex layouts. Further, the generality of our method allows 3D synthesis from a single image. As measured by a comprehensive user study, our method outperforms all existing approaches, preferred by 88-95\%.
\end{abstract}

\addtocontents{toc}{\protect\setcounter{tocdepth}{-1}}

\input{sections/introduction}

\input{sections/shorter_related_work}

\input{sections/shorter_prelim}

\input{sections/method}

\input{sections/results}

\input{sections/applications}
\input{sections/conclusion}

{
    \small
    \bibliographystyle{ieeenat_fullname}
    \bibliography{egbib}
}

\clearpage
\appendix
\twocolumn[{
    \renewcommand\twocolumn[1][]{#1}
    \maketitle
    \vspace{-1.5em}
    \begin{center}
        {\Large \textbf{Supplemental Material:}\\
        \textit{RealmDreamer: Text-Driven 3D Scene Generation\\ with Inpainting and Depth Diffusion}}
    \end{center}
    \vspace{1em}
}]

\addtocontents{toc}{\protect\setcounter{tocdepth}{1}}

\tableofcontents

\input{sections/supplementary}

\end{document}

%% file: include/mymacros.tex
\newcommand{\pname}{RealmDreamer}

\newcommand{\mcal}[1]{\mathcal{#1}}

%% file: sections/introduction.tex
\vspace{-1\baselineskip}
\section{Introduction}
\label{sec:intro}
\begin{figure*}[tp]
    \centering
    \includegraphics[width=\linewidth]{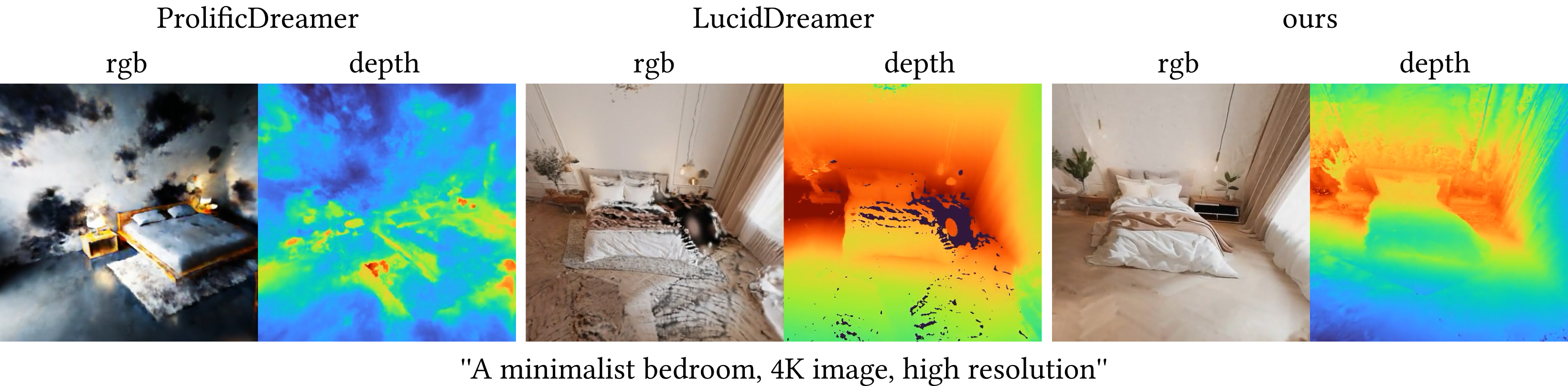}
    \caption{Our method, compared to state-of-the-art ProlificDreamer \cite{wang2023prolificdreamer} and concurrent work LucidDreamer~\cite{luciddreamer}, shows significant improvements. ProlificDreamer yields unsatisfactory geometry and oversaturated renders. LucidDreamer, receiving the same input as our method and an updated depth model~\cite{ke2023repurposing}, displays degeneracy in disoccluded regions, such as the right side of the bed. In contrast, our approach produces visually appealing 3D scenes with realistic geometry. }
    \label{fig:vsd_failure}
\end{figure*}

Text-based 3D scene generation has the potential to revolutionize 3D content creation, with broad applications in virtual reality, game development, and even robotic simulation. However, unlike text-based 2D generative models, 3D data is scarce and lacks diversity, which greatly limits the development of generative 3D techniques. Ideally, one can mitigate this by leveraging rich 2D priors for 3D generation instead. Indeed, object-generation techniques such as DreamFusion~\cite{poole2022dreamfusion} and ProlificDreamer~\cite{wang2023prolificdreamer} do just this, by \textit{distilling} 2D diffusion priors into a 3D representation, with the latter even demonstrating early abilities to generate scenes. Unfortunately, such distillation approaches can often have saturated results, poor geometry, and lack detail, which become very apparent in the more challenging setting of scene generation (\cref{fig:vsd_failure}). This leaves the question: \textit{How to design a distillation technique for high-quality 3D scene generation from pretrained 2D priors?}

A common observation from distillation based object-generation techniques is that greater 3D consistency in 2D diffusion models results in higher-quality distillation, as they provide lower-variance supervision during optimization. As a result, many methods use 2D diffusion models fine-tuned on 3D data~\cite{Deitke2022ObjaverseAU}, such as for novel-view synthesis~\cite{zero123, nerfdiff, magic123}. Equivalent 3D scene datasets are scarce however, which limits the generalization of such techniques to scenes. Alternatively, ProlificDreamer ~\cite{wang2023prolificdreamer} fine-tuned a diffusion model during distillation to be more 3D consistent, producing more highly-detailed textures than before. In this work, we introduce a technique to achieve these strengths \textit{without} requiring 3D training data or fine-tuning existing 2D diffusion models. 

We introduce \textbf{\pname}, a technique for high-fidelity generation of 3D scenes from text prompts (\cref{fig:teaser}). Our key insight is that we can obtain a 3D scene-aware diffusion model for \textit{free}, by simply re-appropriating 2D inpainting diffusion models. Typically, 2D inpainting models condition on a partial image to fill in the rest. Instead, we demonstrate that such models can also condition on a 3D scene and fill in unknown regions for novel view synthesis through our proposed inpainting distillation process. As a result, we obtain high-quality 3D scenes with considerably improved detail and appearance over prior distillation techniques. Further, we propose a simple initialization strategy that provides a 3D scene to use as conditioning for this distillation and serves as an initial point cloud for the 3DGS model. We evaluate our technique on several quantitative metrics and obtain significantly higher quality results than prior work, as notably shown by a user study where we are preferred over state-of-the-art ProlificDreamer~\cite{wang2023prolificdreamer} by 95.5\%. Concretely, our contributions are the following:

 \begin{enumerate}
     \item An occlusion-aware scene initialization for 3DGS, essential for obtaining high-quality scenes (Sec. \ref{subsec:pcd_gen}).
     \item A framework for distillation from 2D inpainting diffusion models which conditions on the existing scene, providing lower variance supervision (Sec.~\ref{subsec:inpainting}).
    \item A method for geometry distillation from diffusion-based depth estimators for higher-fidelity geometry. (Sec.~\ref{subsec:depth}).
     \item State-of-the-art results in text-based generation of 3D scenes, as confirmed by several quantitative metrics and a user study (see Fig.~\ref{fig:result2}, \cref{tab:user_study_results}, \cref{tab:clip_scores}). 
\end{enumerate}

%% file: sections/shorter_related_work.tex
\section{Related Work}
\label{sec:related}

\begin{figure*}[tp!]
    \centering
    \includegraphics[width=\linewidth]{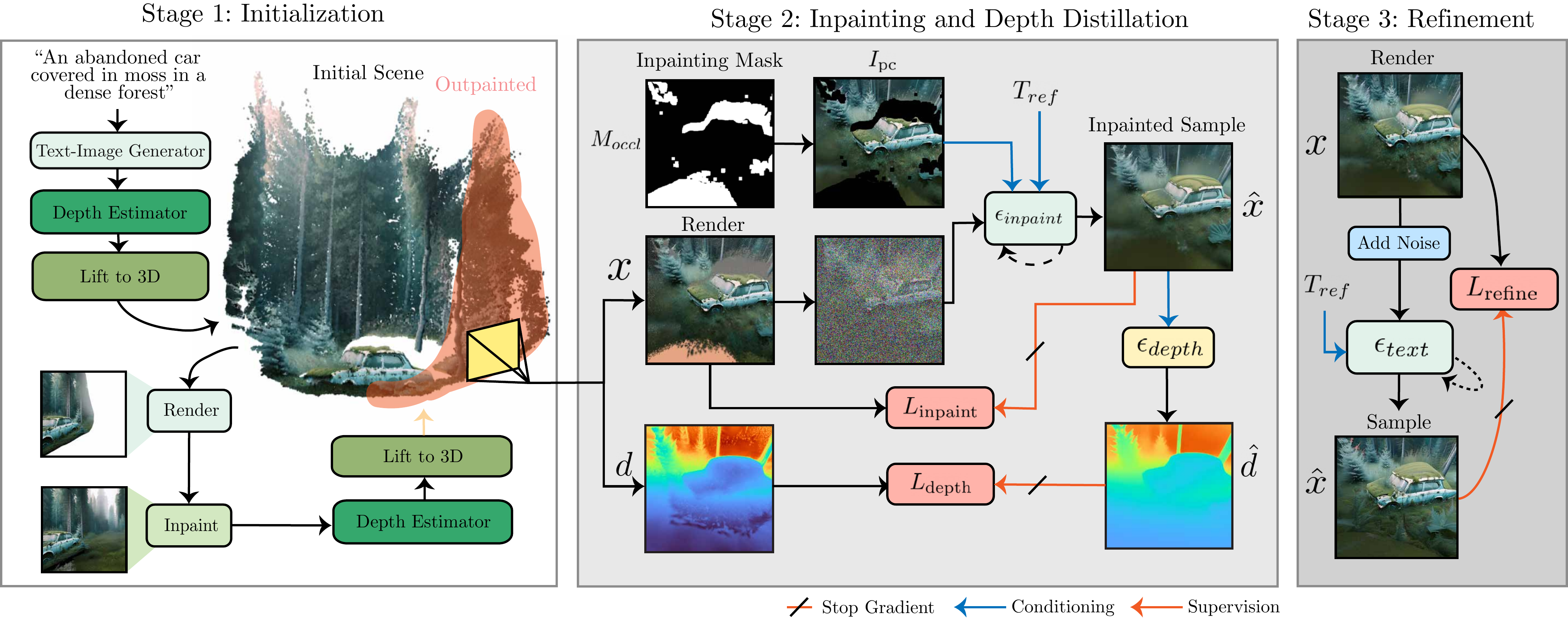}
    \caption{\textbf{Overview of our technique.} Our technique first uses a text prompt and an image to build a point cloud (\cref{subsec:pcd_gen}), which is then completed during the inpainting stage (\cref{subsec:inpainting}) with an additional depth diffusion prior (\cref{subsec:depth}), and finally a refinement stage (\cref{subsec:finetune}) to improve the scene's coherence.}
    \label{fig:method}

\end{figure*}

\noindent\textbf{Text-to-3D.} The first methods for text-to-3D generation were based on retrieval from large databases of 3D assets~\cite{wrodseye, retrieval1, retrieval2}. Subsequently, learning-based methods have dominated ~\cite{text2shape, wgan, liu2023one2345}. However, due to the dearth of diverse paired text and 3D data, many recent methods leverage 2D priors, such as CLIP~\cite{dreamfields, clipforge} or text-to-image diffusion models ~\cite{sjc, poole2022dreamfusion, wang2023prolificdreamer,yi2023gaussiandreamer,gsgen, EnVision2023luciddreamer}. These distill knowledge from 2D priors into a 3D representation, through variations on Dreamfusion's score distillation sampling (SDS) \cite{poole2022dreamfusion}. However, these techniques have primarily been limited to object synthesis. In contrast, there are iterative techniques that incrementally build 3D scenes ~\cite{hoellein2023text2room, luciddreamer} or 3D-consistent perpetual views~\cite{SceneScape}, but can struggle with high parallax. Our proposed technique builds on strengths from distillation and iterative techniques to produce large scale 3D scenes with high parallax using pretrained 2D priors.

\hfill

\noindent\textbf{View Synthesis with Diffusion and 3D inpainting.} Motivated by the success of SDS, several techniques generate 3D objects from a single image by leveraging image-guided diffusion models to generate novel views and distill to 3D~\cite{zhou2023sparsefusion, nerdi}. When trained on larger datasets~\cite{Deitke2022ObjaverseAU}, with better conditioning architectures, these approaches~\cite{mvdream, syncdreamer, zeronvs, zero123, zero123plus} can produce higher quality novel view renders with sharper texture. Some methods also condition denoising directly on renderings from 3D consistent models~\cite{nerfdiff, gennvs} for view synthesis in a multi-view consistent manner. Unfortunately, most techniques rely on object-level data, limiting their use for text-based scene synthesis. 3D inpainting techniques~\cite{spinnerf, diffusion3dinpaint} also leverage image-guided diffusion models to remove small objects in scenes. Other works focus on training custom inpainting models for indoor scenes~\cite{rgbd2} or objects~\cite{invs} to generate novel views. In contrast to these, we leverage pre-trained text-guided inpainting priors and focus on generating large missing regions of diverse scenes with our novel inpainting distillation loss.

\hfill

\noindent\textbf{Concurrent work.}
In the rapidly evolving text-to-3D field, we focus on the most relevant concurrent works, highlighting our key differences. LucidDreamer~\cite{luciddreamer} and Text2NeRF~\cite{text2nerf} uses an iterative approach similar to PixelSynth~\cite{pixelsynth} and Text2Room~\cite{hoellein2023text2room} to generate 3D scenes but displays limited parallax. Considering LucidDreamer as the most relevant concurrent baseline, we compare it in the fairest setting possible, by using newer depth estimators~\cite{ke2023repurposing, depthanything}, and surpass it by $88.5\%$ in our user study. Most recently, in follow-up work, CAT3D~\cite{cat3d}, utilizes a diffusion model finetuned on multiview datasets to generate multiple views from a single image. In contrast, our entire pipeline does not use multiview images.

%% file: sections/shorter_prelim.tex
\section{Preliminaries}
\label{sec:prelim}

\subsection{3D Gaussian Splatting}
\label{subsec:gs}
3D Gaussian Splatting (3DGS)~\cite{3dgs} has recently emerged as an explicit alternative to NeRF \cite{mildenhall2020nerf}, offering extremely fast rendering speeds and a memory-efficient backwards pass. In 3DGS, a set of splats are optimized from a set of posed images. The soft geometry of each splat is represented by a mean $\mu\in\mathbb{R}^3$, scale vector $s\in\mathbb{R}^3$, and rotation $R$ parameterized by quaternion $q\in\mathbb{R}^4$, so that the covariance of the Gaussian is given by $\Sigma=RSS^TR^T$ where $S=\text{Diag}(s)$. Additionally, each splat has a corresponding opacity $\sigma\in\mathbb{R}$ and color $c\in\mathbb{R}^3$. 

The splats $\{\Theta_i\}_{i=1}^N=\{\mu_i, s_i,q_i,\sigma_i, c_i\}_{i=1}^N$ are projected to the image plane where their contribution $\alpha_i$ is computed from the projected Gaussian (see \cite{zwick}) and $\sigma_i$. A pixel's color is obtained by $\alpha$-blending Gaussians sorted by depth:
\begin{equation}
\label{eq:rendering}
C = \sum_{i=1}^N \alpha_i c_i\prod_{j=1}^{i-1} (1 - \alpha_j).
\end{equation}

A significant drawback of 3DGS-based approaches is the necessity of a good initialization. State-of-the-art results are only achieved with means $\mu_i$ initialized by the sparse depth of Structure-from-Motion~\cite{sfm}, which is not applicable for scene generation. To address this challenge, we generate a prototype of our 3D scene using a text prompt, which we then optimize (\cref{subsec:pcd_gen}).

\begin{figure*}[tp]
    \centering
    \includegraphics[width=0.9\linewidth]{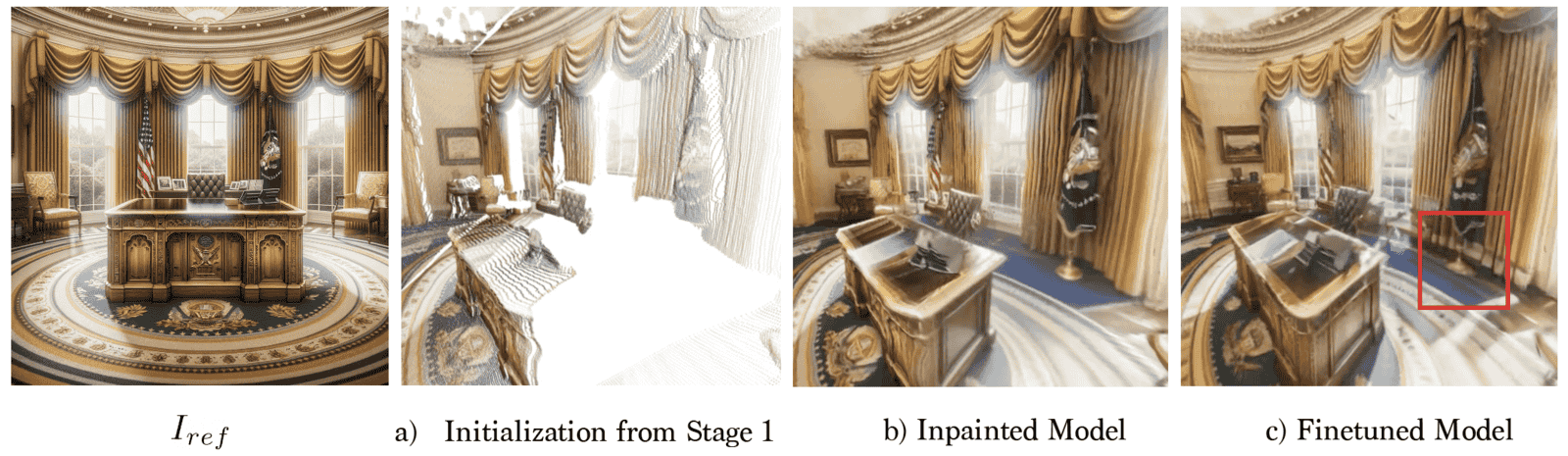}
    \caption{\textbf{Progression of 3D Model after each stage.} We show how the 3D model changes after each stage in our pipeline. As shown in a) Stage 1 (\cref{subsec:pcd_gen}) creates a point cloud with many empty regions. In b), we show the subsequent inpainted model from Stage 2 (\cref{subsec:inpainting}). Finally, the fine-tuning stage (\cref{subsec:finetune}) refines b) to produce the final model, with greater cohesion and sharper detail.}
    \label{fig:progression}
\end{figure*}

\subsection{Conditional Diffusion Models}
\label{subsec:sds}

Diffusion models~\cite{thermo, ddpm, edm, ddim, song1,song2} are generative models which learn to map noise $x_T\sim\mcal{N}(0, I)$ to data by iteratively denoising a set of latents $x_t$ corresponding to decreasing noise levels $t$ using non-deterministic DDPM~\cite{ddpm} or deterministic DDIM sampling~\cite{ddim}, among others~\cite{edm, song1,song2}.

Given $t$, a diffusion model $\epsilon_\theta$ is trained to predict the noise $\epsilon$ added to the image such that we obtain $\epsilon_\theta(x_t, t)$, which approximates the direction to a higher probability density. Often, the data distribution is conditional on quantities such as text $T$ and images $I$, so the denoiser takes the form $\epsilon_\theta(x_t,I,T)$. In the conditional case, classifier-free guidance is often used to obtain the predicted noise~\cite{cfg, pix2pix}: 
\begin{equation}
\begin{aligned}
\tilde{e}_{\theta}(x_t, I, T) =&~~e_{\theta}(x_t, \emptyset, \emptyset) \\&
+ S_I \cdot (e_{\theta}(x_t, I, \emptyset) - e_{\theta}(x_t, \emptyset, \emptyset)) \\
&+ S_T \cdot (e_{\theta}(x_t, I, T) - e_{\theta}(z_t, I, \emptyset)) 
\end{aligned}
\end{equation}
where $\emptyset$ indicates no conditioning, and the values $S_I$ and $S_T$ are the guidance weights for image and text, dictating fidelity towards the respective conditions. In the case of latent diffusion models like Stable Diffusion~\cite{sd}, denoising happens in a compressed latent space by encoding and decoding images with an encoder $\mcal{E}$ and decoder $\mathsf{D}$.

\textbf{Score Distillation Sampling.} Distilling text-to-image diffusion models for text-to-3D generation of object-level data has enjoyed great success since the introduction of Score Distillation Sampling (SDS)~\cite{poole2022dreamfusion, sjc}. Given a text prompt $T$ and a text-conditioned denoiser $\epsilon_\theta(x_t,T)$, SDS optimizes a 3D model by denoising noised renderings. Given a rendering from a 3D model $x$, we sample a timestep and corresponding $x_t$. Considering $\hat x = \frac 1 {\alpha_t} (x_t - \sigma_t \epsilon_\theta(x_t,T))$ as the detached one-step prediction of the denoiser, SDS is equivalent to minimizing~\cite{hifa}: 
\begin{equation}
L_\text{sds}=\mathbb{E}_{t, \epsilon} \left[ w(t)\left\| x - \hat x  \right\|_2^2 \right]
\end{equation}
where $w(t)$ is a time-dependent weight over all cameras with respect to the parameters of the 3D representation, and the distribution of $t$ determines the strength of added noise. In this work, we use a variation of SDS to distill from pretrained-inpainting models (\cref{subsec:inpainting})

%% file: sections/method.tex
\section{Method}
\label{sec:method}

We now describe our technique in detail, which broadly consists of three stages: \textbf{initialization} (left of \cref{fig:method}, \cref{subsec:pcd_gen}); \textbf{inpainting} (middle of \cref{fig:method}, \cref{subsec:inpainting}) with \textbf{depth distillation} (middle of \cref{fig:method}, \cref{subsec:depth}); and \textbf{finetuning} (right of \cref{fig:method}, \cref{subsec:finetune}). Given a text-prompt $T_{\text{ref}}$ and camera poses, we initialize the scene-level 3DGS representation $\{\Theta_i\}_{i=1}^N$ leveraging 2D diffusion models and monocular depth priors, along with the computed \emph{occlusion volume} (\cref{subsec:pcd_gen}). With this robust initialization, we use 2D inpainting models to predict novel views, distilling to 3D to create a complete 3D scene (\cref{subsec:inpainting}). In this stage, we also incorporate depth distillation for higher-quality geometry (\cref{subsec:depth}). Finally, we refine the model with a sharpness filter on sampled images to obtain high-quality 3D samples (\cref{subsec:finetune}). The result from these stages are shown in \cref{fig:progression}.

\begin{figure*}[tp]
    \centering
    \includegraphics[width=0.95\textwidth]{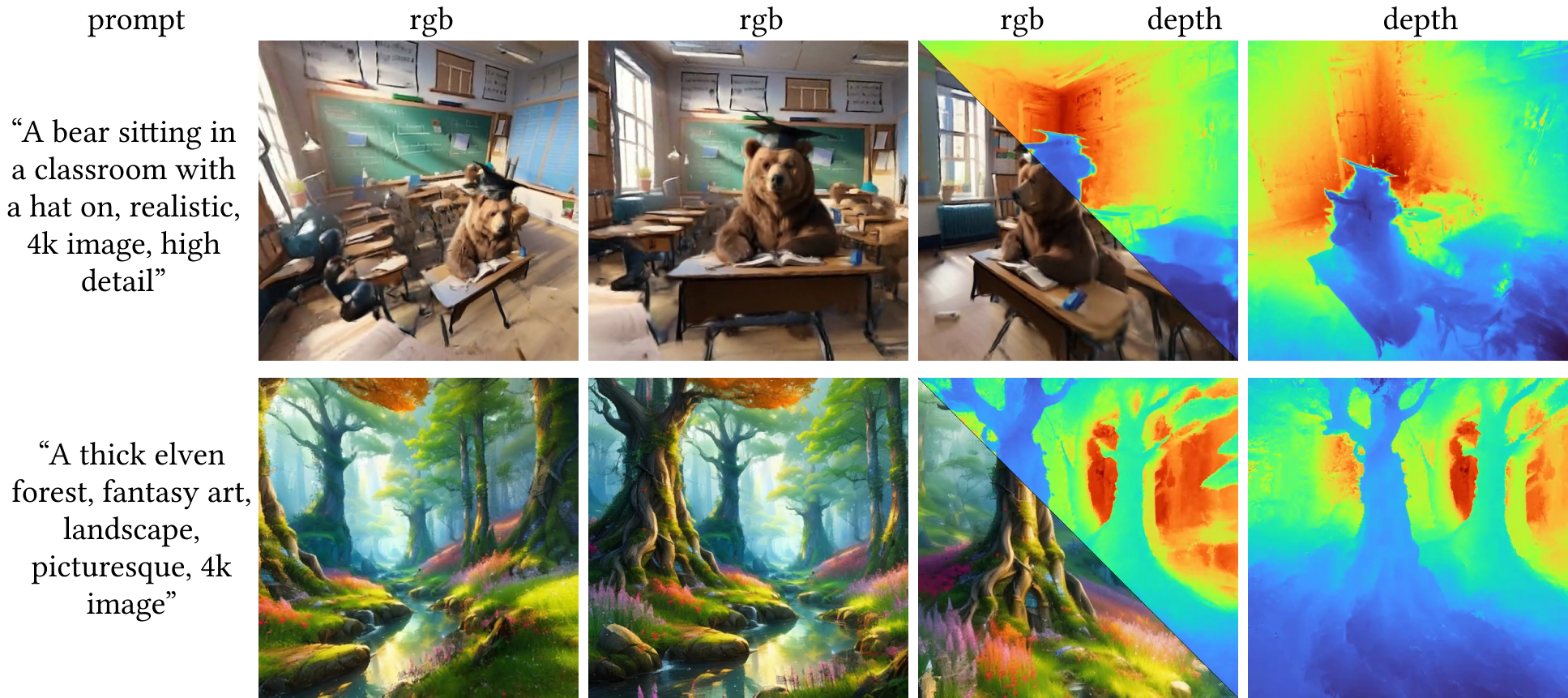}
    \caption{\textbf{Qualitative Results.} In the left column, we show the input prompt for our technique. In the next two columns, we show the renderings from our 3D model from different viewpoints. In the fourth column, we show the level of agreement between rendering and geometry by a split view of the rendering and depth. Finally, in the last column, we show the depth map.}
    \label{fig:result}
\end{figure*}

\subsection{Initializing a Scene-level 3D Representation}
\label{subsec:pcd_gen}

Our technique utilizes 3DGS for text-conditioned optimization, making a good initialization essential. A common strategy in this setting is to initialize with a sphere \cite{poole2022dreamfusion, lin2023magic3d} but the density of a scene is more complex and distributed. Hence, we leverage pretrained 2D priors to synthesize a robust initialization (left of \cref{fig:method}).

Concretely, we first generate a reference image of the scene $I_{ref}$ from the text prompt $T_{ref}$ with a state-of-the-art text-to-image-model. We then employ a monocular depth model~\cite{ke2023repurposing} $\mcal{D}$ to lift this image to a pointcloud $\mcal{P}$ from corresponding camera pose $P_{ref}$. Depending on the generated image, the extent of the pointcloud can vary widely. To make the initialization more robust, we \textit{outpaint} $I_{ref}$ by moving the camera left and right of $P_{ref}$ to poses $P_{aux}$. We use an inpainting diffusion model~\cite{sd} to fill in the unseen regions which are lifted to 3D using $\mcal{D}$. The union of all generated points thus becomes $\mcal{P}$. 

\textbf{Determining Incomplete Regions.} Given the initial point cloud $\mathcal{P}$, we then precompute the undetermined 3D region, or the \emph{occlusion volume} $\mathcal{O}$, which is the set of voxel centers within the scene's occupancy grid which are occluded by the existing points in $\mathcal{P}$ from $P_{ref}$. We use $\mcal{O}$ when computing inpainting masks later and define the initialization of our 3DGS means as
\begin{equation}
\label{eq:init}
    \{\mu_i\}_{i=1}^N = \mathcal{P} \cup \mathcal{O}.
\end{equation}
More details can be found in the supplementary.

\subsection{Inpainting Diffusion for 3D-Conditioned Distillation}
\label{subsec:inpainting}

Since our initialization is generated from sparse poses, viewing it from novel viewpoints exposes large holes in disoccluded regions (\cref{fig:progression}). We resolve this with a novel inpainting distillation technique, that conditions a 2D inpainting diffusion model $\epsilon_{\text{inpaint}}$~\cite{sd} on the existing scene to complete missing regions. The model takes as input a noisy rendering $x_t$ of $\{\Theta_i\}_{i=1}^N$, and is conditioned by the text prompt $T_{\text{ref}}$, an occlusion mask $M_\text{occl}$, and the point cloud render $I_{\text{pc}}$. Sampling from this model results in novel views $\hat x$ which plausibly fill in the holes in the renderings while preserving the structure of the 3D scene (\cref{fig:method}).

\textbf{Conditioning the inpainting model.} To compute the conditioning mask $M_\text{occl}$ for $\epsilon_{\text{inpaint}}$, we render the point cloud $\mathcal{P}$ and the precomputed occlusion volume $\mathcal{O}$. We set all components of $M_\text{occl}$ for which the occlusion volume is visible from the target to $0$, and $1$ otherwise. Note that this handles cases such as the point cloud occluding itself (see the supplement for a visualization).

\textbf{Computing the inpainting loss.} Our 2D inpainting diffusion model $\epsilon_{\text{inpaint}}$~\cite{sd} operates in latent space, thus additionally parametrized by its encoder $\mcal{E}$ and decoder $\mathsf{D}$. We render an image $x$ with the initialized 3DGS model, and encode it to obtain a latent $z$, where $z=\mcal{E}(x)$. We then add noise to this latent, yielding $z_t$, corresponding to a randomly sampled timestep $t$ from the diffusion model's noise schedule. Using these quantities, we take multiple DDIM~\cite{ddim} steps from $z_t$ to compute a clean latent $\hat z$ corresponding to the inpainted image. 

\begin{figure*}[tp]
    \centering
    \includegraphics[width=0.95\textwidth]{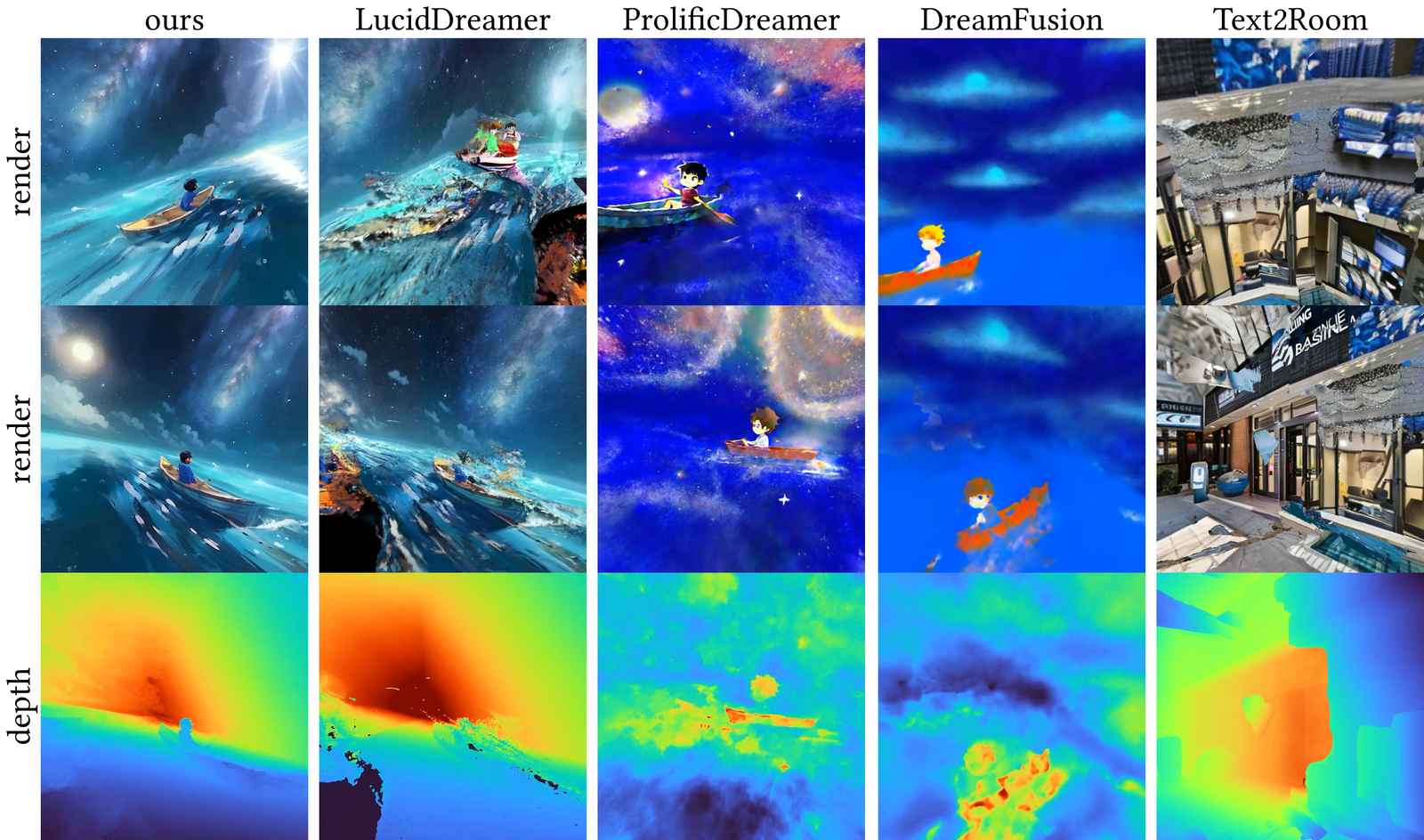}
    \caption{\textbf{Qualitative Comparisons.} Our technique shows superior quality in appearance and geometry than all baselines. Please see the supplementary for more comparisons. Prompt: ``A boy sitting in a boat in the middle of the ocean, under the milkyway, anime style". }
    \label{fig:result2}
\end{figure*}

We define our inpainting loss in both latent space and image space, by additionally decoding the predicted latent to obtain $\hat{x} = \mathsf{D}(\hat{z})$. We compute the L2 loss between the latents of the render and sample, as well as an L2 and LPIPS perceptual~\cite{lpips} loss between the rendered image and the decoded sample. To prevent edits outside of the inpainted region, we also add an anchor loss on the unmasked region of $x$, as the L2 difference between $x$ and original point cloud render $I_{pc}$. Our final inpainting loss is
\begin{equation}
\begin{aligned}
    \label{eq:inpaint_loss}
    L&\vphantom{}_\text{inpaint} = \lambda_{\text{latent}} ||z - \hat z||_2^2 +  \lambda_{\text{image}} ||x - \hat x||_2^2 \\ &+ \lambda_{\text{lpips}} \text{LPIPS}(x, \hat x) +\lambda_{\text{anchor}} ||M_\text{occl}(x - I_{\text{pc}})||_2^2
\end{aligned}
\end{equation}
with $\lambda$ weighting the different terms. We discuss the similarity of this loss with SDS in the supplemetary.
\\

\textit{Discussion}. In contrast to existing iterative methods which utilize inpainting (such as Text2Room and LucidDreamer), our framework does not iteratively construct a scene with inpainting. In practice, sampling from inpainting models often produces artifacts (such as due to out-of-distribution masks), which iterative approaches can amplify when generating from new poses. In contrast, due to scene-conditioned multiview optimization, we obtain cohesive 3D scenes and do not progressively accumulate errors. Moreover, in contrast to DreamFusion and ProlificDreamer, our method utilizes a scene-conditional diffusion model, providing lower variance updates for effective optimization (see row 2 of \cref{fig:ablation}). This avoids the high-saturation and blurry results that are typically found (\cref{fig:result2}).

\subsection{Depth Diffusion for Geometry Distillation}
\label{subsec:depth}

To improve the quality of generated geometry, we incorporate a pretrained geometric prior to avoid degenerate solutions. Here, we leverage monocular depth diffusion models and propose an additional depth distillation loss (middle of \cref{fig:method}). Crucially, we integrate this with our inpainting distillation by conditioning the depth model $\epsilon_{\text{depth}}$ on the aforementioned samples $\hat{x}$ from  $\epsilon_{\text{inpaint}}$.

Our insight is that these samples $\hat{x}$ act as suitable, in-domain, conditioning for the depth diffusion model throughout optimization, while renders $x$ can be incoherent before convergence. Further, this ensures that predictions from $\epsilon_{\text{depth}}$ are aligned with $\epsilon_{\text{inpaint}}$ despite not using a RGBD prior. Starting from pure noise $d_1\sim\mathcal{N}(0,I)$, we predict the normalized depth using DDIM sampling \cite{ddim}. We then compute the (negated) Pearson Correlation between the rendered depth and sampled depth:

\begin{equation}
    L_\text{depth}= -\frac{\sum (d_i - \frac{1}{n}\sum d_k)(\hat{d}_i - \frac{1}{n}\sum \hat{d}_k)}{\sqrt{\sum (d_i - \frac{1}{n}\sum d_k)^2 \sum (\hat{d}_i - \frac{1}{n}\sum \hat{d}_k)^2}}
\end{equation}
where $d$ is the rendered depth and $n$ is the number of pixels.

\subsection{Optimization and Refinement}
\label{subsec:finetune}

The final loss for the first training stage of our pipeline is thus:
\begin{equation}
    L_\text{init} = L_\text{inpaint} + L_\text{depth}.
 \end{equation}
After training with this loss, we have a 3D scene that roughly corresponds to the text prompt, but which may lack cohesiveness between the reference image $I_{ref}$ and the inpainted regions (see \cref{fig:progression}). To remedy this, we incorporate an additional lightweight refinement phase. In this phase, we utilize a vanilla text-to-image diffusion model $\epsilon_\text{text}$ personalized for the input image with Dreambooth~\cite{dreambooth, dreambooth3d, nerdi, realfusion}. We compute $\hat x$ using the same procedure as in \cref{subsec:inpainting}, except with $\epsilon_\text{text}$. The loss $L_\text{text}$ is the same as \cref{eq:inpaint_loss}, except with the $\hat z$ and $\hat{x}$ sampled with this finetuned diffusion model $\epsilon_\text{text}$. Note that the noise added to the renderings at this stage is smaller to combat the higher variance samples from the lack of image conditioning.

 We also propose a novel sharpening procedure: instead of using $\hat{x}$ to compute the image-space diffusion loss introduced earlier, we use $\mcal{S}(\hat{x})$, where $\mcal{S}$ is a sharpening filter applied on samples from the diffusion model. Finally, to encourage high opacity points in our 3DGS model, we incorporate an opacity loss $L_{\text{opacity}}$ per point that encourages a point's opacity to reach either 0 or 1, inspired by the transmittance regularizer used in Plenoxels \cite{plenoxels}. The combined loss for the fine-tuning stage is:

\begin{equation}
    L_\text{refine} = L_\text{text} + \lambda_\text{opacity} L_\text{opacity},
\end{equation}
where $\lambda_{\text{opacity}}$ controls the effect of the opacity loss.

\subsection{Implementation Details}

\textbf{Point Cloud Initialization.} We implement this stage (\cref{subsec:pcd_gen}) in Pytorch3D \cite{ravi2020pytorch3d}, with Stable Diffusion \cite{sd} for outpainting. To lift the generated images to 3D, we use Marigold~\cite{ke2023repurposing}, a monocular depth estimation model. Since it predicts relative depth, we align its predictions with the metric depth predicted by DepthAnything \cite{depthanything}.

\noindent \textbf{Inpainting and Refinement Stage.} Our inpainting (\cref{subsec:inpainting}) and refinement stages (\cref{subsec:finetune}) are implemented in NeRFStudio \cite{nerfstudio} using the official implementation of Gaussian Splatting \cite{3dgs}. We use Stable Diffusion 2.0 as $\epsilon_{\text{text}}$ and its inpainting variant as $\epsilon_\text{inpaint}$, building on threestudio \cite{threestudio2023} to define our diffusion-guided losses. Further, we use Marigold \cite{ke2023repurposing} as our depth diffusion model. During the inpainting stage, we set the guidance weight for image and text conditioning of $\epsilon_\text{inpaint}$ as 1.8 and 7.5 respectively, and sample the timestep $t$ from $\mathcal{U}(0.1, 0.95)$. We find that a high image guidance weight produces samples with greater overall cohesion. We also use a guidance weight of 7.5 for the text-to-image diffusion model $\epsilon_\text{text}$ during the refinement stage, sampling noise from $\mathcal{U}(0.1, 0.3)$.

\noindent \textbf{Timing.} The first stage, currently unoptimized, takes 2.5 hours. The inpainting stage, trained for 15,000 iterations, runs for 8 hours on a 24GB Nvidia A10 GPU. The refinement stage, at 3,000 iterations, completes in 2.5 hours on the same GPU.

%% file: sections/results.tex
\section{Results}

\input{sections/dataset}
\label{sec:results}

We evaluate our technique on a custom dataset of 20 prompts, and associated camera poses $P_i$, selected to showcase parallax and disocclusion. We built this dataset by creating a set of 20 prompts, and having a human expert manually choose camera poses using a web-viewer \cite{nerfstudio}, by displaying a scene prototype obtained as in \cref{subsec:pcd_gen}. No such dataset already exists for this problem, as existing text-to-3D techniques~\cite{poole2022dreamfusion, wang2023prolificdreamer} typically operate with spherical camera priors. Please refer to the supplemental video results to see the generated scenes.

\subsection{Qualitative Results}

We show some qualitative results in~\cref{fig:result} with additional results in the supplementary, demonstrating effective 3D scene synthesis across various settings (indoor, outdoor) and image styles (realistic, fantasy, illustration). We would like to highlight the rendering quality and the consistency of rendering and geometry, underscoring our method's use of inpainting and depth priors.

\begin{figure}[tp]
    \centering
    \includegraphics[width=\linewidth]{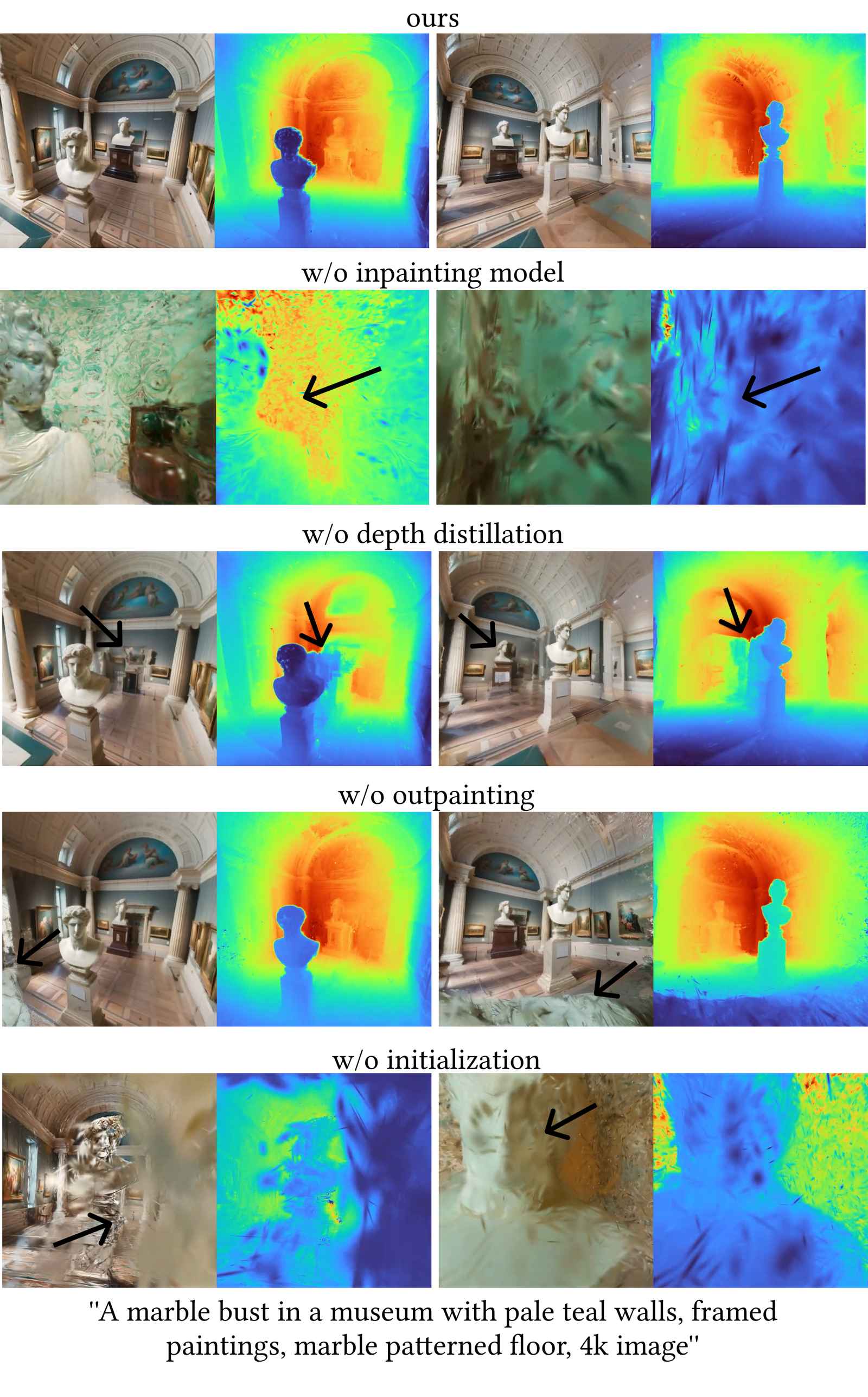}
    \caption{\textbf{Ablation Results.} We show the qualitative results of our model and its ablations. Arrows indicate failures in the ablated models. Please see ~\cref{subsec:ablations} for a detailed discussion of the ablated components and their respective importance.}
    \label{fig:ablation}
\end{figure}

\subsection{Comparisons}

We compare our technique with state-of-the-art for text-to-3D that use either distillation or iterative approaches: DreamFusion~\cite{poole2022dreamfusion}, ProlificDreamer~\cite{wang2023prolificdreamer}, Text2Room~\cite{hoellein2023text2room}, and concurrent work LucidDreamer~\cite{luciddreamer} (\cref{fig:result2}). Both ProlificDreamer and DreamFusion generate oversaturated scenes with incorrect geometry and scene structure. On the other hand, Text2Room fails to construct non-room scenes, as it deviates from the input prompt during generation. Similarly, LucidDreamer's~\cite{luciddreamer} scenes lack cohesion, with noisy results in occluded regions. Note that LucidDreamer and Text2Room take an image as input; we gave these baselines the same input image as to ours and updated their depth models.

\subsection{User Study}
\label{sec:user_study}

To validate the quality of our generated 3D scenes, we conduct a user study~(\cref{tab:user_study_results}), similar to prior work~\cite{wang2023prolificdreamer, magic3d, fantasia}. We conducted a study on Amazon Mechanical Turk and recruited 20 participants with a `master' qualification to compare each baseline, while following several guidelines outlined in \cite{Bylinskii2022TowardsBU}. The details can be found in the supplementary. Participants overwhelmingly prefer results from our technique over baselines.

\begin{table}[h!]
\centering
\caption{\textbf{Results of user study.} We show the percentage of comparisons where our technique was preferred over baselines: PD~\cite{wang2023prolificdreamer}, DF~\cite{poole2022dreamfusion}, T2R~\cite{hoellein2023text2room}, and LD~\cite{luciddreamer}.}
\begin{tabular}{l@{\hspace{0.5em}}c@{\hspace{0.5em}}c@{\hspace{0.5em}}c@{\hspace{0.5em}}c}
\hline
& Ours vs. PD & Ours vs. DF & Ours vs. T2R & Ours vs. LD \\
\hline
& \textbf{95.5\%} & \textbf{94.5\%} & \textbf{88\%} & \textbf{88.5\%} \\
\hline
\end{tabular}
\label{tab:user_study_results}
\vspace{-0.1cm}
\begin{tabular}{@{}p{8cm}@{}}
\end{tabular}
\end{table}

\begin{table}[ht]
\caption{\textbf{CLIP alignment scores and additional metrics} for scene renderings of our method and the baselines. CLIP scores are scaled by 100. Higher is better for all metrics.}
\centering
\begin{tabular}{l c c c}
\hline
Method & CLIP & Depth Pearson & IS \\ \hline
Ours & \textbf{31.69} & \textbf{0.89} & \textbf{6.99} \\
Text2Room~\cite{hoellein2023text2room} & 28.11 & 0.77 & 5.10 \\
DreamFusion~\cite{poole2022dreamfusion} & 29.48 & 0.09 & 6.80 \\
ProlificDreamer~\cite{wang2023prolificdreamer} & 29.39 & 0.16 & 6.89 \\
LucidDreamer~\cite{luciddreamer} & 29.97 & 0.80 & 5.73 \\ \hline
\end{tabular}
\label{tab:clip_scores}
\end{table}

\subsection{Quantitative Metrics}
\label{sec:quant}

We also provide a quantitative comparisons with all baselines based on alignment to the text prompt using CLIP~\cite{clip}, Inception Score~\cite{inception_score} on renderings, and the quality of geometry with the pearson correlation between rendered depth and the predicted depth by DepthAnythingV2~\cite{depth_anything_v2}. We note that due to the lack of ground truth data, standard reconstruction metrics such as PSNR or LPIPS~\cite{lpips} do not apply. We compute these scores for renderings from the same trajectory and the corresponding prompt for all scenes. As Text2Room's results degrade significantly away from the initial pose, we compare with a render from the initial pose for CLIP. As shown in \cref{tab:clip_scores}, our method shows significantly better performance across all metrics.

\subsection{Ablations}
\label{subsec:ablations}
We verify the proposed contributions of our method by ablating the key components in ~\cref{fig:ablation} with the specified prompt (\cref{tab:ablation_study}). In the first row, we show our method. In the second row, we show the importance of the low variance samples from the inpainting diffusion model (\cref{subsec:inpainting}). Distillation with a vanilla text-to-image model as in the final stage, results in high-variance samples causing the 3DGS representation to diverge. In the third row, we remove $L_\text{depth}$; this results in incorrect geometry and incoherent renderings. Note in particular the discrepancy in the background when viewing from left versus right. In the fourth row, we initialize our method using only the reference image $I_{ref}$ without outpainting at the neighbouring poses $P_\text{aux}$. This results in poor results in the corresponding regions, as they lack a good initialization. Finally, in the last row, we show our result without using the $\mu$ initialization from \cref{eq:init}, which results in divergence.

\begin{table}[ht]
\caption{\textbf{Ablation Study Results} showing the impact of different components on Depth Pearson correlation and CLIP score. CLIP scores are scaled by 100. Higher is better for both metrics.}
\centering
\begin{tabularx}{\linewidth}{ 
  >{\raggedright\arraybackslash}X %
  >{\centering\arraybackslash}X %
  >{\centering\arraybackslash}X %
}
\hline
Ablation & Depth & CLIP \\ \hline
No Depth Loss & 0.86 & 31.55 \\
No Initialization & 0.42 & 20.31 \\
No Inpainting & 0.50 & 21.14 \\
No Outpainting & 0.79 & 31.00 \\
Ours & \textbf{0.90} & \textbf{33.10} \\ \hline
\end{tabularx}
\label{tab:ablation_study}
\end{table}

\subsection{Application: Single image to 3D}
\label{sec:single_image}

\begin{figure}
    \centering
    \includegraphics[width=\linewidth]{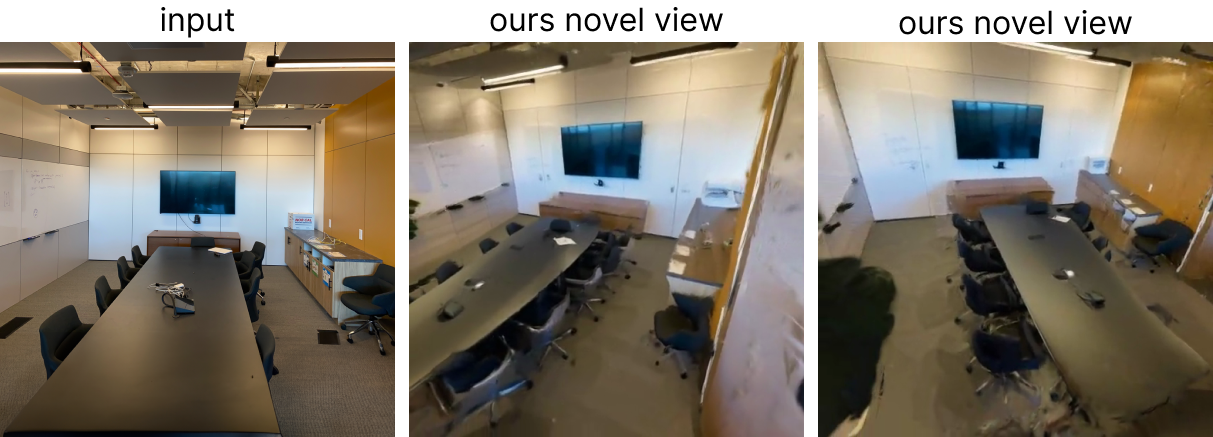}
    \caption{\textbf{Result for single-image to 3D.} Using a provided image and a prompt obtained via an image captioning model, our technique can generate a 3D scene and fill in occluded regions.}
    \label{fig:single_image}
\end{figure}

Our technique extends to creating 3D scenes from a single image, as shown in \cref{fig:single_image}, by using a user's image as $I_{ref}$ and a text-prompt $T_{ref}$ obtained using an image-captioning model. Our pipeline can effectively fill in occluded areas and generate realistic geometry for unseen regions.

%% file: sections/conclusion.tex
\section{Conclusion}
\label{sec:conclusion}

We have proposed \textbf{\pname}, a method for generation of forward-facing 3DGS scenes leveraging inpainting and depth diffusion. Our key insight was to leverage the lower variance of image conditioned (inpainting) diffusion models for synthesis of 3D scenes, providing much higher quality results than existing baselines as measured by a comprehensive user study. Still, limitations remain; our method takes several hours, and produces blurry results for complex scenes with significant disocclusion. Future work may explore efficient diffusion models for faster training, and conditioning for 360-degree generations.

\section{Acknowledgements}
We thank Jiatao Gu and Kai-En Lin for early discussions, Aleksander Holynski and Ben Poole for later discussions. This work was supported in part by an NSF graduate Fellowship, ONR grant
N00014-23-1-2526, NSF CHASE-CI Grants 2100237 and 2120019, gifts from
Adobe, Google, Qualcomm, Meta, the Ronald L. Graham Chair, and the UC
San Diego Center for Visual Computing.

%% file: sections/supplementary.tex
\begin{figure}[bp!]
    \centering
    \includegraphics[width=\linewidth]{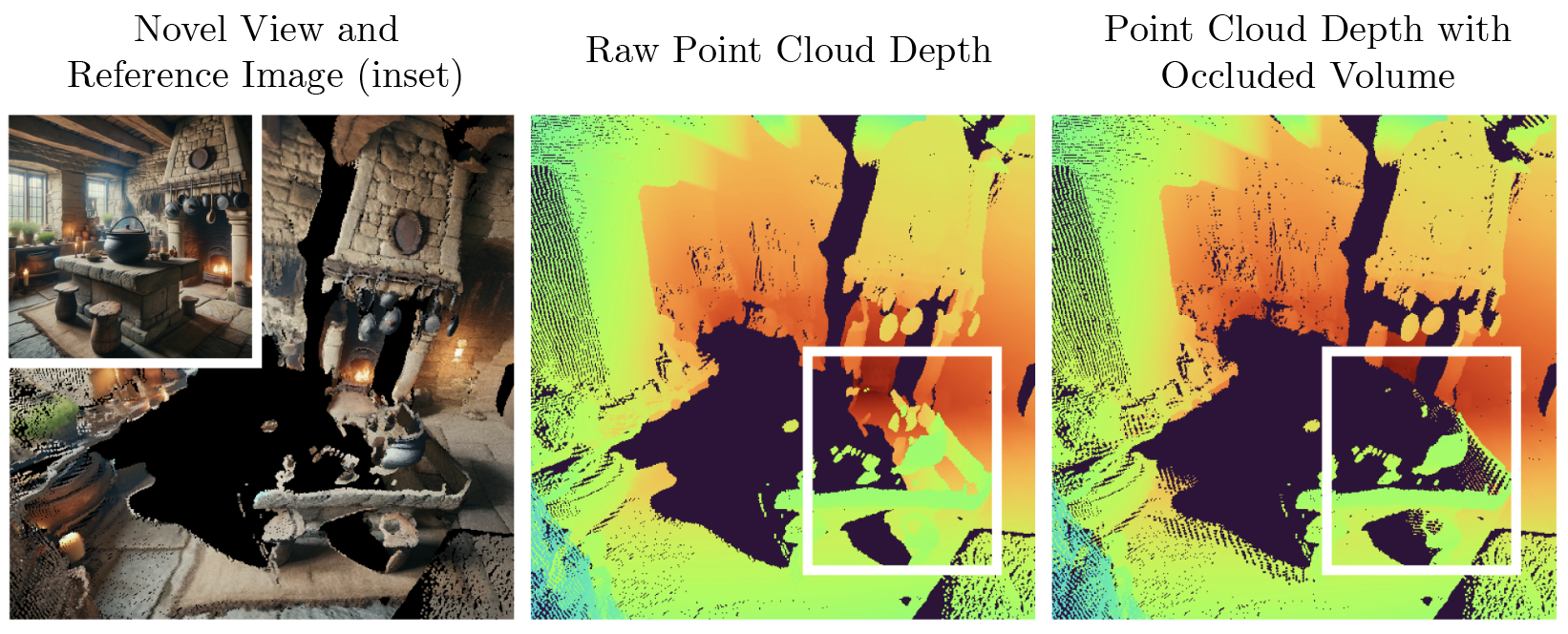}
    \caption{\textbf{Creating the Inpainting Mask.} 3D inpainting requires filling in a 3D volume, which is not always equivalent to missing regions in 2D point cloud renders. By computing an \textit{occlusion volume}, we avoid situations where the floor is visible through the table (middle) but instead could be occluded. The right depth map accounts for the ambiguity of the volume in the table. }
    \label{fig:occl}
\end{figure}

\section*{Ethical Considerations}

While we use pretrained models for all components of our pipeline, it is important to acknowledge biases and ethical issues that stem from the training of these large-scale image generative models \cite{sd}. As these models are often trained on vast collections of internet data, they can reflect negative biases and stereotypes against certain populations, as well as infringe on the copyright of artists and other creatives. It is essential to consider these factors when using these models and our technique broadly. 

\section{Why is the occlusion volume important?}

Our key contribution is the inpainting distillation loss that provides lower variance and high-quality supervision for text-to-3D, compared to regular text-to-image model based distillation, as shown in the ablations. Given that we use 2D inpainting models for 3D inpainting via this distillation, we must ask: \textit{how to compute the 3D region that needs to be inpainted?} 

We proposed a simple technique to do so by computing the \textbf{occluded volume} (described in~\cref{subsec:occl_comp}), which is the 3D region occluded by objects in the reference image $I_{\text{ref}}$. Note that regardless of what objects may be present in this occluded region, rendering from $P_{\text{ref}}$ would yield $I_{\text{ref}}$, as elements in the image would occlude the objects. The 2D inpainting masks we obtain then must hence, reflect this unknown 3D region, as that is the part of the scene left to complete.

Instead of computing the occlusion volume, another alternative is using the holes from point cloud renderings as the inpainting mask. Indeed, these holes also represent unknown 3D regions. However, the 3D region indicated by such 2D masks is a \textbf{subset} of the occluded 3D region and hence, incomplete. This is shown in~\cref{fig:occl}, which shows the masked point cloud depth, where the masked region represents the inpainting mask. When using the holes in the point cloud as an inpainting mask, one can observe that the back side of the kitchen table is visible. In reality, a solid kitchen table would never expose this face. In contrast, using the occlusion volume, we can correctly determine the entire 3D region missing in $I_{\text{ref}}$, which can be visually verified by comparing images. Specifically, the latter takes into account self-occlusion, providing a more accurate estimate and allowing details to fill into the occluded region. 

In practice, such self-occlusions do not appear for all prompts yet is important to maintain the correctness of the 3D inpainting formulation. If there are many renderings such as in~\cref{fig:occl}, the quality of inpainted samples would be incorrect and lead to a noisier distillation process.

\section{Discussion on Baselines}

We are among the first to tackle text-to-3D scenes and showcase a high-level of parallax. As a result, there are limited open-source 3D scene generation techniques to compare with. We choose to compare with techniques that either use distillation - a key part of our pipeline and iterative approaches - which shares similarity with the initialization technique we use.

\subsection{Comparison with Dreamfusion and ProlificDreamer} By comparing with prior SOTA distillation techniques \cite{poole2022dreamfusion, wang2023prolificdreamer}, we demonstrate that these approaches are suboptimal for scene generation, which has not been demonstrated before. Arguably, distillation techniques should be able to build any 3D scene since the 2D priors used are general, assuming object-centric regularizers and prompts are absent. However, we empirically find that simple distillation from text-to-image models is insufficient for wide baselines. This comparison highlights the importance of our inpainting distillation, which conditions on a partial 3D scene, enabling high level of parallax not seen in prior work. 

We also note that ProlificDreamer~\cite{wang2023prolificdreamer} first showcased scene-level results using distillation, indicating that distillation is not purely for object-generation. Still, it presented a limited set of scenes and did not show high parallax as it focused on rotating a camera through a scene. Our baseline comparison attempts to test the performance of this approach on wide camera trajectories and finds that it often produces hazy results. 

\subsection{Relation between SDS and our distillation}

Our distillation is similar to the score distillation loss (SDS) used in Dreamfusion~\cite{poole2022dreamfusion}. However, unlike prior work, we do not use just text conditioning, but also renders from the point cloud. As a result, the classifier-free guidance weight we is much lower - at 7.5, avoiding over-saturated results typically found with SDS. Further, unlike SDS which denoises noisy renders in one step, we take multiple steps with DDIM sampling. Further, we also use a loss on the denoised latent and decoded image, to produce high quality supervision.

\subsection{Comparison with LucidDreamer and Text2Room} Our initialization step is a key part of the pipeline and reminiscent of prior and concurrent iterative techniques which incrementally grow a scene. Yet, such techniques have yet to showcase high quality over high-parallax camera trajectories, which we demonstrate. We find that incremental generation of 3D scenes can lead to noise accumulation, due to errors in monocular depth and alignment of geometry. Hence, unlike concurrent work LucidDreamer~\cite{luciddreamer}, we do not limit our scene generation to our initialization step. We rely on our inpainting distillation loss to produce highly cohesive 3D scenes, by distilling across multiple views, rather than building a scene incrementally. We also note that while LucidDreamer does use 3DGS optimization, it is a more conventional reconstruction-based optimization, with no inpainting or geometry priors incorporated. As a result, its optimization stage is very different from our inpainting distillation, which provides rich priors for appearance and geometry at every iteration.

Further, we also avoid many limitations of prior work such as Text2Room, which often produces scenes with low-prompt alignment, especially those in outdoor scenes. We attribute this to the technique deleting regions of the mesh before inpainting, as the original technique prescribes. When such deletions accumulate over time, they are prone to erasing parts of the original scene defined in $I_{\text{ref}}$ entirely. In contrast to this, our technique maintains a simple initialization strategy and relies on a high-quality inpainting distillation process to fill-in mission regions, without sacrificing the quality of the initialized regions.

\subsection{Implementation}

\textbf{Text2Room \cite{hoellein2023text2room} and LucidDreamer \cite{luciddreamer}} We use the official implementation of Text2Room and LucidDreamer on Github. To ensure a fair comparison, we estimate depth using Marigold~\cite{ke2023repurposing} and DepthAnything~\cite{depthanything} as in our technique, replacing the original IronDepth~\cite{irondepth} and ZoeDepth~\cite{zoedepth} respectively. The rest of the pipeline is kept the same.

\textbf{ProlificDreamer~\cite{wang2023prolificdreamer} and Dreamfusion~\cite{poole2022dreamfusion}} We use the implementation of these baselines provided in threestudio~\cite{threestudio2023} and use their recommended parameters, training for 25k and 10k steps, respectively. To ensure a fair comparison, we use the same poses for these baselines as our technique.

\section{Additional ablations and discussion}
\label{subsec:add_ablation}
\subsection{Use of DDIM Inversion.} During the inpainting and refinement stage (Sec 4.2, 4.3 in the original paper), we find it helpful to obtain the noisy latent $z_t$ using DDIM inversion \cite{ddim}, where $z=\mcal{E}(x)$, $x$ is the rendered image, and $t$ is a timestep corresponding to the amount of noise added. This is similar to prior work on 2D/3D editing and synthesis using pre-trained diffusion models \cite{hertz2022prompt, EnVision2023luciddreamer}. We demonstrate the importance of doing so in \cref{fig:ddim_inversion}, where DDIM inversion can significantly improve the detail in the optimized model. During the inpainting stage, we use 25 steps to sample an image from pure noise, and during refinement, we use 100 steps.

\subsection{Use of sharpening filter.} 
\begin{figure}[t!]
    \centering
    \includegraphics[width=\linewidth]{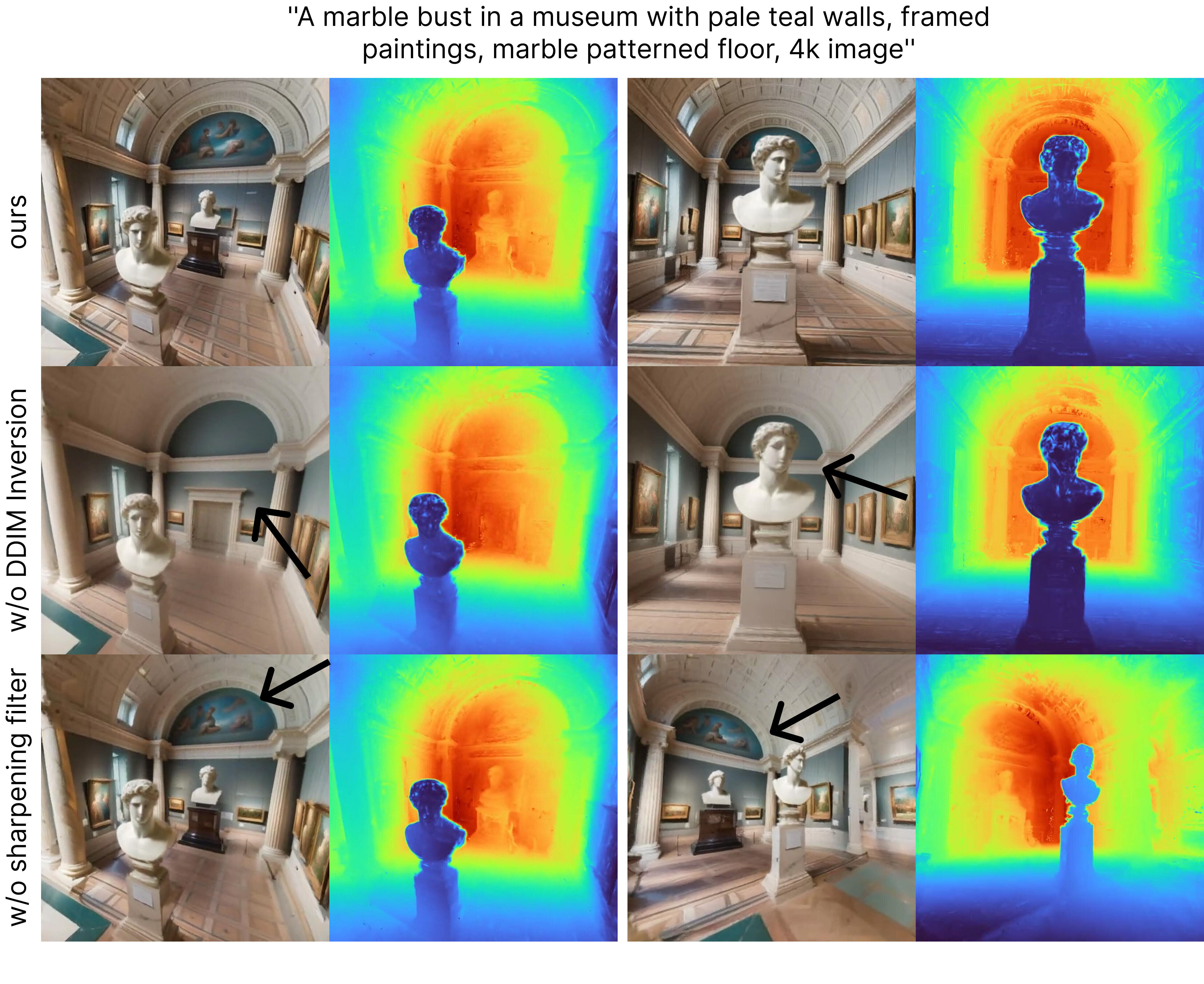}
    \caption{We ablate the importance of DDIM Inversion and applying a sharpening filter. As in ~\cite{EnVision2023luciddreamer}, we find that DDIM inversion allows more details to be synthesized by our method. Additionally, we find that detail slighly increases when applying a sharpening filter to the sampled images.}
    \label{fig:ddim_inversion}
\end{figure}

In \cref{fig:ddim_inversion}, we also see that applying a sharpening filter to the sampled images results in slightly more detail. We attribute this to the blurry nature of some samples of the diffusion model.

\section{Additional Implementation Details}
\label{sec:implementation_suppl}

We intend to open-source our code upon publication. In addition, we describe some key implementation details to assist reproducibility.

\subsection{Point Cloud Generation}

\textbf{Image Generation.} We generate our reference image $I_\text{ref}$ using a variety of state-of-the-art text-image generation models, choosing between Stable Diffusion XL~\cite{sdxl}, Adobe Firefly, and DALLE-3~\cite{dalle3}. 

\textbf{Depth Estimation.} As mentioned earlier, we use Marigold~\cite{ke2023repurposing} as our depth estimation model, with absolute depth obtained using DepthAnything~\cite{depthanything}. We align the relative depth with this absolute depth by computing the linear translation that minimizes the least squares error between them. Since DepthAnything provides separate model weights for indoor and outdoor scenes, we use GPT-4 to decide which checkpoint to use by passing $I_{ref}$ as input. When iteratively growing the point cloud, we follow Text2Room \cite{hoellein2023text2room} and align the predicted depth with the ground truth depth rendered via Pytorch3D~\cite{ravi2020pytorch3d} for all regions with valid geometry. We additionally blur the edges of these regions to lower the appearance of seams at this intersection. 

\textbf{Growing the pointcloud beyond $P_{ref}$.} After lifting the reference image $I_{ref}$ to a pointcloud $\mcal{P}$, we additionally create new points from neighbouring poses $P_{aux}$, as mentioned earlier. In practice, we notice that using the same prompt $P_{ref}$ across all neighbouring poses $P_{aux}$ can lead to poor results, as objects mentioned in the prompt get repeated. Hence, we use GPT-4 to compute a new suitable prompt that can represent the neighbouring views of $P_{ref}$. Specifically, we pass the reference image $I_{ref}$, the original prompt $T_{ref}$ and ask GPT-4 to provide a new prompt $T_{aux}$ that can be suitable for neighbouring regions. For instance, when viewing a \textit{"car in a dense forest"}, $T_{aux}$ may correspond to a \textit{"dense forest"}.

\subsection{Occlusion Volume Computation}
\label{subsec:occl_comp}

We compute the occlusion volume $\mcal{O}$ with Bresenham's line-drawing algorithm. First, we initialize an occupancy grid $\mcal{G}$ using the point cloud $\mcal{P}$ from stage 1. We also store whether any voxel is occluded with respect to $P_{ref}$ within the same occupancy grid, initially settings all voxels as occluded.
Then, we draw a line from the position of the reference camera $T_{ref}$ to all voxels in the occupancy grid $\mcal{G}$, iterating over the voxels covered by this line and marking all as \textit{non-occluded} until we encounter an occupied voxel. Once the algorithm terminates, all voxels that are untouched by the line-drawing algorithm form our occlusion volume $\mcal{O}$.

\subsection{Optimization}

\textbf{Hyperparameter Weights.} We set $\lambda_\text{latent} = 0.1, \lambda_\text{anchor}=10000$ during the inpainting stage, and $\lambda_\text{latent}=0.01, \lambda_\text{anchor}=0$ during the refinement stage. The other parameters are set as $\lambda_\text{image}=0.01, \lambda_\text{lpips}=100, \lambda_\text{depth}=1000\text{, and } \lambda_\text{opacity}=10$.

\textbf{Use of Dreambooth during fine-tuning.} While fine-tuning the output from stage 2, we use Dreambooth \cite{dreambooth} to personalize the text-to-image diffusion model with the reference image $I_{ref}$ and associated prompt $T_{ref}$. We find that this helps the final 3D model adhere closer to $I_{ref}$ stylistically. We use the implementation of Dreambooth from HuggingFace and train at a resolution of 512x512 with a batch size of 2, with a learning rate of 1e-6 for 200 steps.

\textbf{Opacity Loss.} We compute the opacity loss as the binary cross entropy of each splat's opacity $\sigma_i$ with itself. This encourages the opacity to reach either 0 or 1.

\textbf{Gaussian Splatting.} We initialize our gaussian splatting model during the inpainting stage, using the point cloud from stage 1, where each point is an isotropic gaussian, with the scale set based on the distance to its nearest neighbors. During the inpainting stage, we use a constant learning rate of $0.01$ for rotation, $0.001$ for the color, and $0.01$ for opacity. The learning rate of the geometry follows an exponentially decaying scheduler, which decays to $0.00005$ from $0.01$ over $100000$ steps, after $5000$ warmup steps. Similarly, the scale is decayed to 0.0001 from $0.005$ over $10000$ steps, after $7000$ warmup steps. During the refinement stage, we use a constant learning rate of $0.01$ for rotation, $0.001$ for the color, $0.01$ opacity, and $0.0001$ for scale. We use an exponentially decaying scheduler for the geometry, which decays to $0.0000005$ from $0.0001$ over $3000$ steps, after $750$ warmup steps. During the inpainting distillation, we also dilate $M_{\text{occl}}$ to improve cohesion at mask boundaries. Further, we find it essential to mask the latent-space L2 loss, to prevent unwanted gradients outside the masked region.

\begin{figure*}[t]
    \centering
    \includegraphics[width=\linewidth]{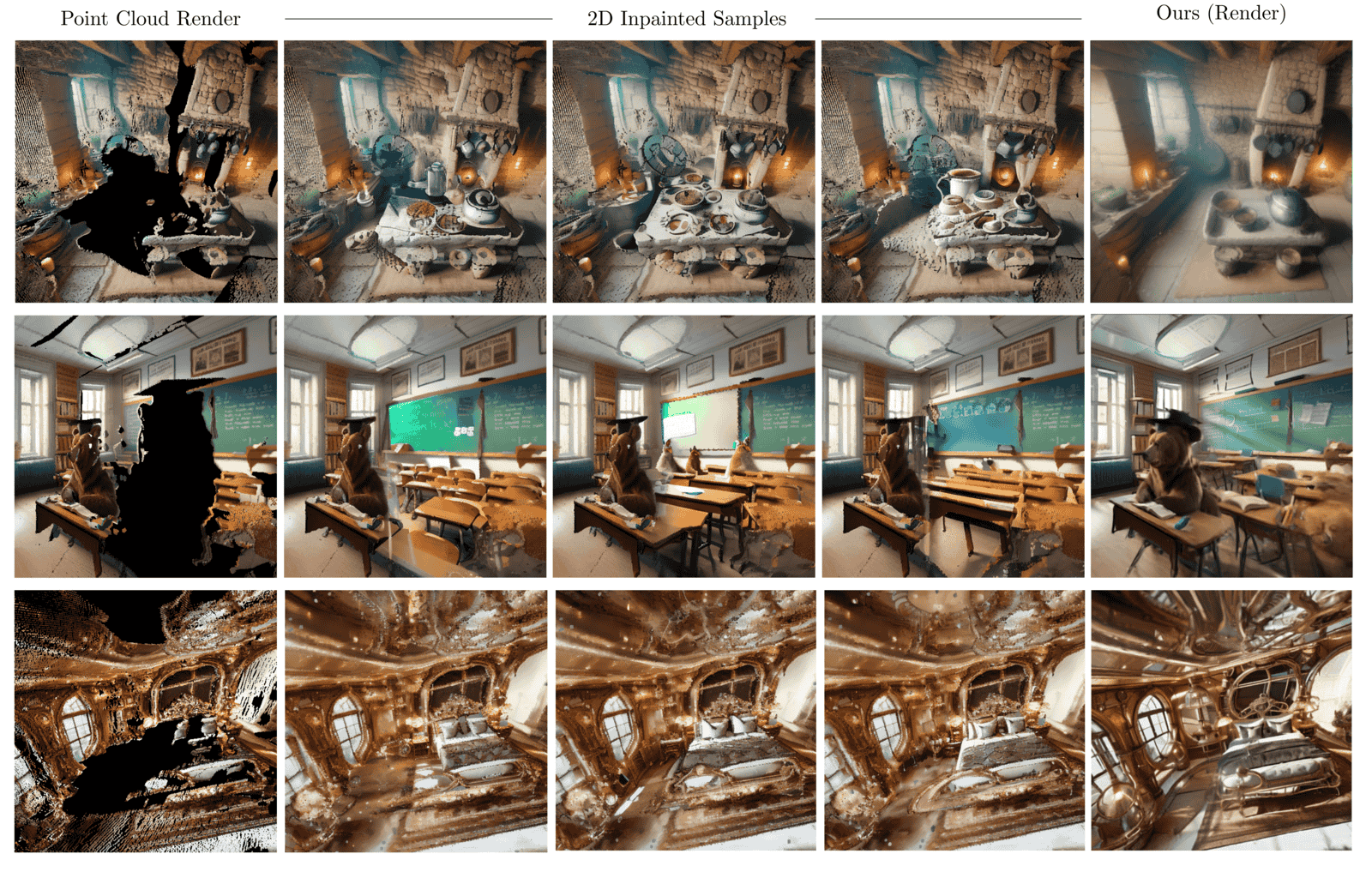}
    \caption{\textbf{Comparison of sampling from 2D inpainting models and our optimized model.} Left: Renders from the point cloud generated in stage 1. Middle (cols 2-4): Inpainted Samples of the previous render using an occlusion-based inpainting mask and Stable Diffusion~\cite{sd}. Right: A render from our final 3DGS model for the corresponding scene. We find that our distillation techniques produce results with high cohesion while avoiding many artifacts from ancestral sampling of 2D inpainting models.}
    \label{fig:inpainting_samples}
\end{figure*}

\section{User Study}
\label{subsec:user_study_details}

For comparison with ProlificDreamer~\cite{wang2023prolificdreamer}, DreamFusion~\cite{poole2022dreamfusion}, and LucidDreamer~\cite{luciddreamer}, we showed participants side-by-side videos comparing our method to the baseline. For fairness, we use the same camera trajectory in all videos. The order of the videos was also randomized to prevent any biases due to the order of presentation. The user's preference was logged along with a brief explanation.

When comparing with Text2Room~\cite{hoellein2023text2room}, instead of a video, we showed users side-by-side sets of three multiview images for each prompt, due to the degeneracy of the output mesh far from the starting camera pose. The user's preferred triplet was logged along with their brief explanation. The images we showed looked slightly left and right of the reference pose $P_\text{ref}$.

\subsection{Common themes of the user study.}

All study participants were asked to justify their preferences for one 3D scene over the other after making their choice. Participants were not informed about the names or the nature of any technique. We also adopted method-neutral language to avoid biasing the user to prefer any particular technique. We find that their provided reasoning closely aligns with several noted limitations of the baselines, which we discuss further:

\textbf{ProlificDreamer~\cite{wang2023prolificdreamer} can produce cloudy results}. Several participants described the NeRF renders as containing ``moving clouds'', a ``hazy atmosphere'', and a ``blotch of colours''. This can likely be attributed to the presence of floaters in the model, which is evident in the noisy depth maps shown in \cref{sec:qual_suppl}. In contrast, participants described our method as ``clean and crisp when it comes to the colors and sharpness of the pixels'' and looking realistic, without the presence of over-saturated colors.

\textbf{Dreamfusion~\cite{poole2022dreamfusion} lacks realism and detail.} Feedback from users when comparing with DreamFusion often mirrored feedback from the ProlificDreamer comparison, referencing a lack of realism and detail in the produced renders. One participant said ``[Our technique] is more crisp and does a better job with the content quality.'', while the Dreamfusion result can ``feel disjointed''. Another participant described a render as having a ``distorted looking background''. In contrast to these issues, our technique synthesizes realistic models with high detail and high-quality backgrounds, with minimal blurriness.

\textbf{Text2Room~\cite{hoellein2023text2room} can produce messy outputs.} A common theme across feedback regarding Text2Room was that it often looked like a mess, sometimes with a ``strange distortion''. One user writes that our result is ``less busy and fits the description''. Another common reason users cited when choosing our technique was the adherence to the input prompt, with Text2Room often missing key objects that are expected for an associated prompt. Our technique, however, is capable of producing highly coherent outputs that are faithful to the reference prompt and produce high-quality renderings from multiple views.

\textbf{LucidDreamer~\cite{luciddreamer}'s scenes lack cohesion and can be distorted.} Multiple participants pointed out that LucidDreamer's scenes degrade in quality when moving away from the initial pose. One participant wrote ``The image on the left loses cohesion when rotated." referring to LucidDreamer and in contrast another wrote ``There is less visual distortion when the camera is moved around the room." about \pname. Some participants also noted that objects produced by our technique were more \textit{solid}, with one participant noting ``The shapes are solid on the right and hold their form.". These comments underscore the limitations of purely iterative approaches.

\section{Additional Discussion}
\label{subsec:discu_addition}

\subsection{Impact of Distillation}

In \cref{fig:inpainting_samples}, we show the importance of our distillation process for filling in occluded regions and the challenge in doing so. Column 1 shows renders following stage 1, which contains large holes, giving objects a \textit{thin} look (such as the bear in row 2 or the table in row 1). By computing an occlusion volume and obtaining inpainting masks, we can inpaint these renders to obtain several inpainted samples (columns 2-4). However, these samples can contain several artifacts. For instance, in row 1 of \cref{fig:inpainting_samples}, the surface of the table is quite cluttered in individual samples. This is likely due to the challenge of inpainting images with complex masks that are out of distribution. These images also show the challenge in building cohesive scenes with single view inpainting. For instance the blackboard in row 2 has multiple shades of green in the 2D samples. Despite these challenges, our final render for the scene, in column 5 of \cref{fig:inpainting_samples} is clean and free of stray artifacts such as bright colours or ambiguous objects. We attribute this difference to our distillation process. 

As mentioned earlier, since we optimize over multiple views, we are less susceptible to artifacts present in individual samples and can produce 3D inpaintings that satisfy multiple views. Prior work, such as Text2Room~\cite{hoellein2023text2room} instead relies primarily on dilating masks and deleting regions of generated scenes to simplify the inpainting process. Our inpainting distillation process does not require any aggressive modification to the scene but can produce high-quality results. We highly encourage the viewer to view the video renderings to appreciate the extent of occluded regions that our distillation technique generates.

\section{Limitations}

\begin{figure}[t]
    \centering
    \includegraphics[width=\linewidth]{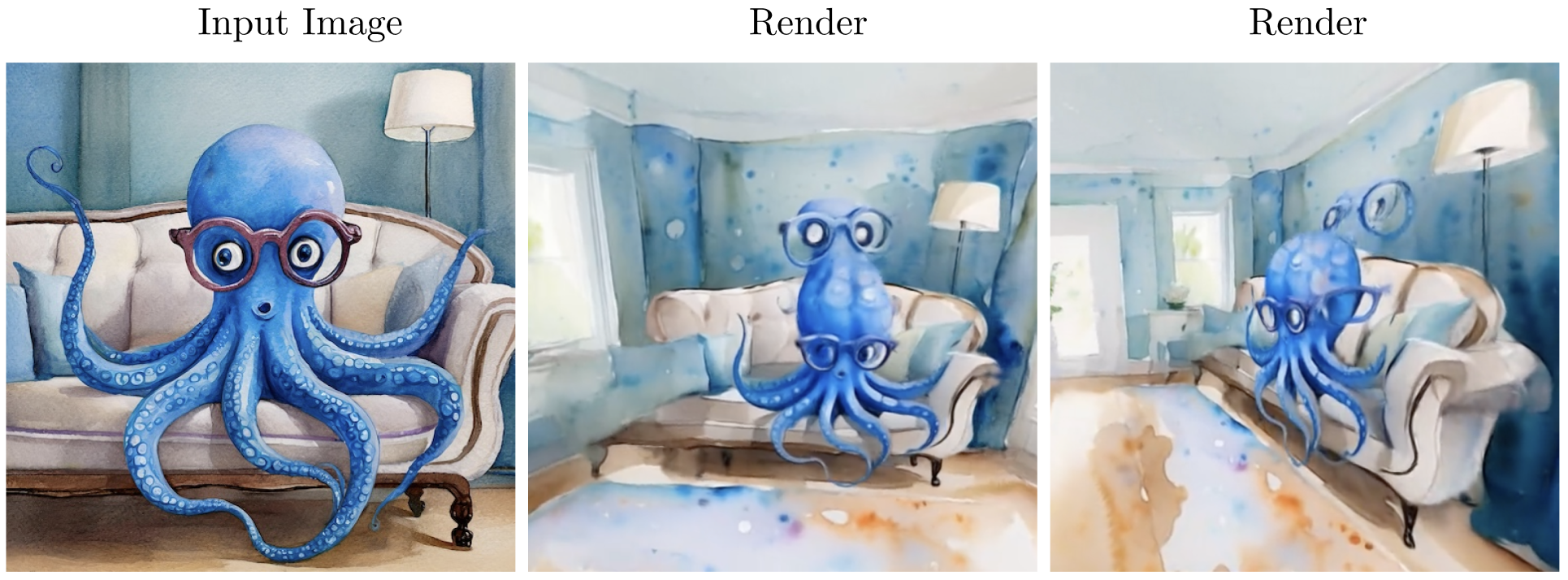}
    \caption{\textbf{Janus Problem due to multi-view optimization.} Since we optimize over multiple-views, sometimes the final model can show the same object multiple times to satisfy all views, such as the pair of glasses above the octopus. Prompt: ``A blue octopus wearing glasses on a couch in the living room, watercolor style" }
    \label{fig:janus}
\end{figure}

\textbf{Janus Problem.} By adopting a distillation based approach, we occasionally encounter the Janus problem, where the face of an object appears multiple times across renders. An example is shown in \cref{fig:janus} As we focus on scene generation and additionally condition on the 3D scene, this is less pronounced than in object generation~\cite{poole2022dreamfusion} and can likely be alleviated with view-dependent prompting~\cite{armandpour2023re}.

\textbf{Artifacts in rendering.} Some scenes also display artifacts at the surface of objects over a wide baseline. We believe improvements to our 3DGS implementation, such as by incorporating anti-aliasing, and surface regularizers might help with this. We note that our results are still significantly better than prior work and uses only 2D priors.

\section{Additional Qualitative Results}
\label{sec:qual_suppl}

In the following pages, we show qualitative results from our technique as well as all baselines.

\clearpage
\thispagestyle{empty}
\begin{figure*}
    \centering
    \includegraphics[height=1\textheight,keepaspectratio]{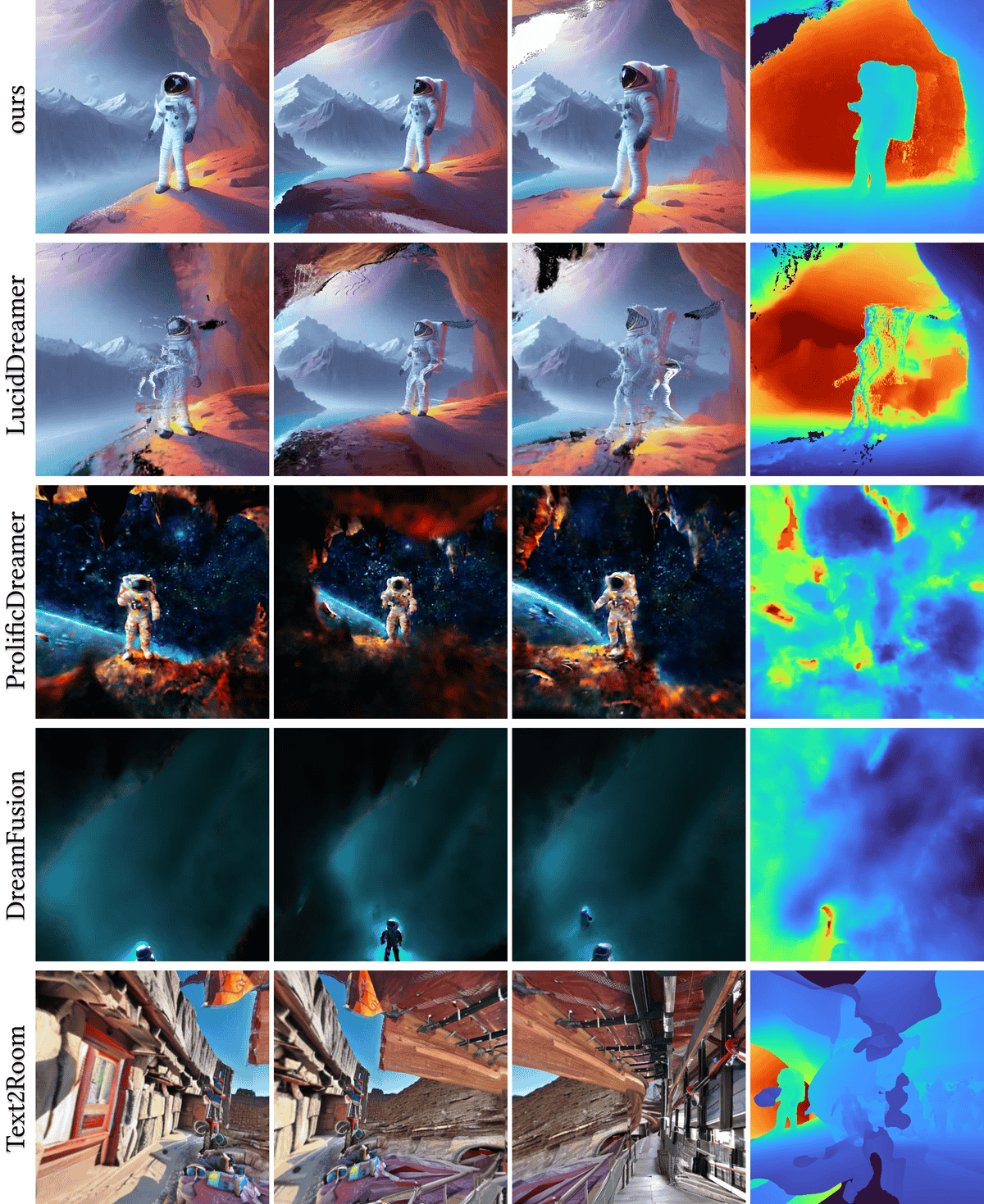}
    \captionsetup{labelformat=empty}
    \caption{\huge Prompt: "An astronaut in a cave, trending on artstation, 8k image"}
\end{figure*}

\clearpage
\thispagestyle{empty}
\begin{figure*}
    \centering
    \includegraphics[height=1\textheight,keepaspectratio]{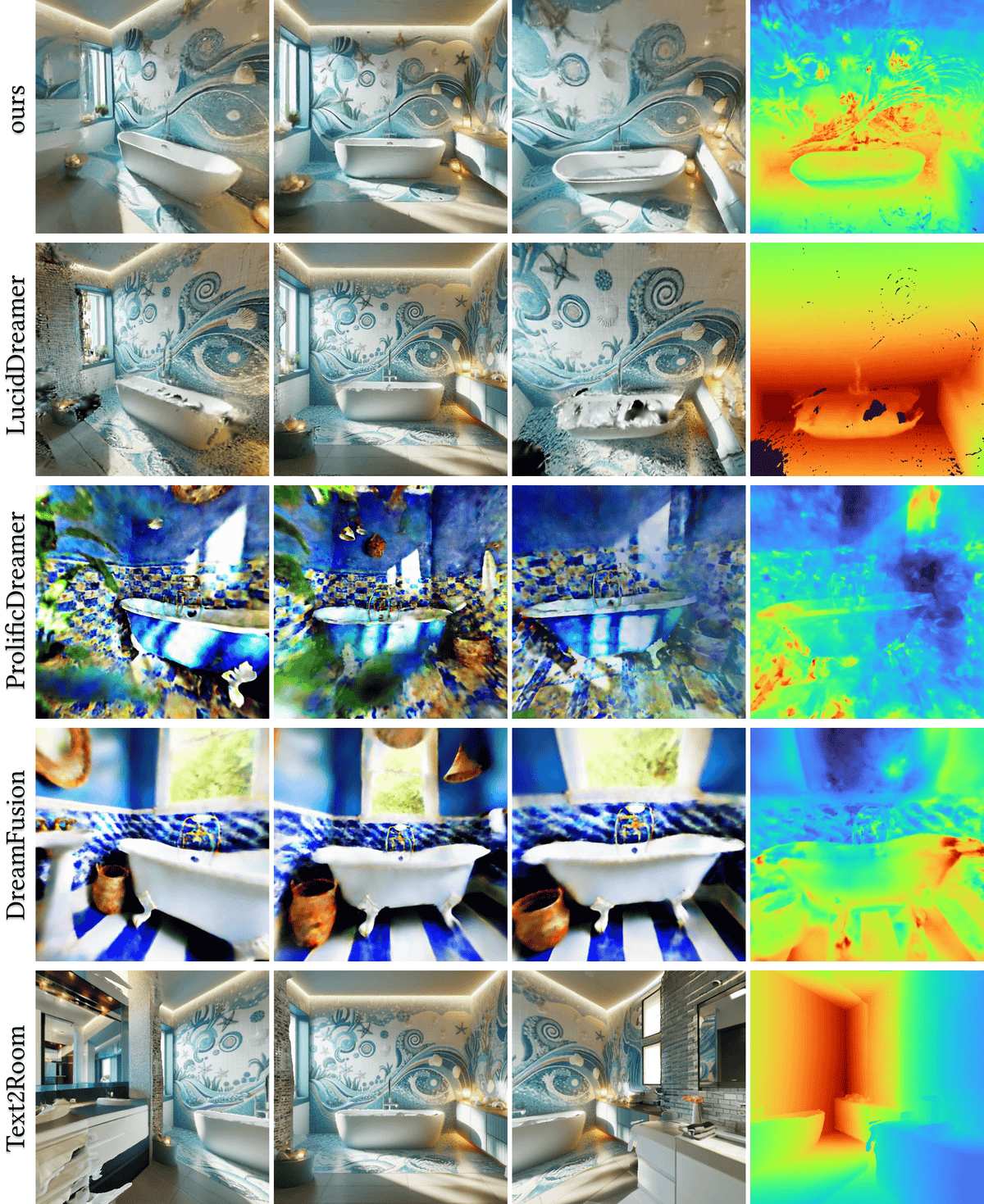}
    \captionsetup{labelformat=empty}
    \caption{\huge Prompt: "Editorial Style Photo, Coastal Bathroom, Clawfoot Tub, Seashell, Wicker, Mosaic Tile, Blue and White"}
\end{figure*}

\clearpage
\thispagestyle{empty}
\begin{figure*}
    \centering
    \includegraphics[height=1\textheight,keepaspectratio]{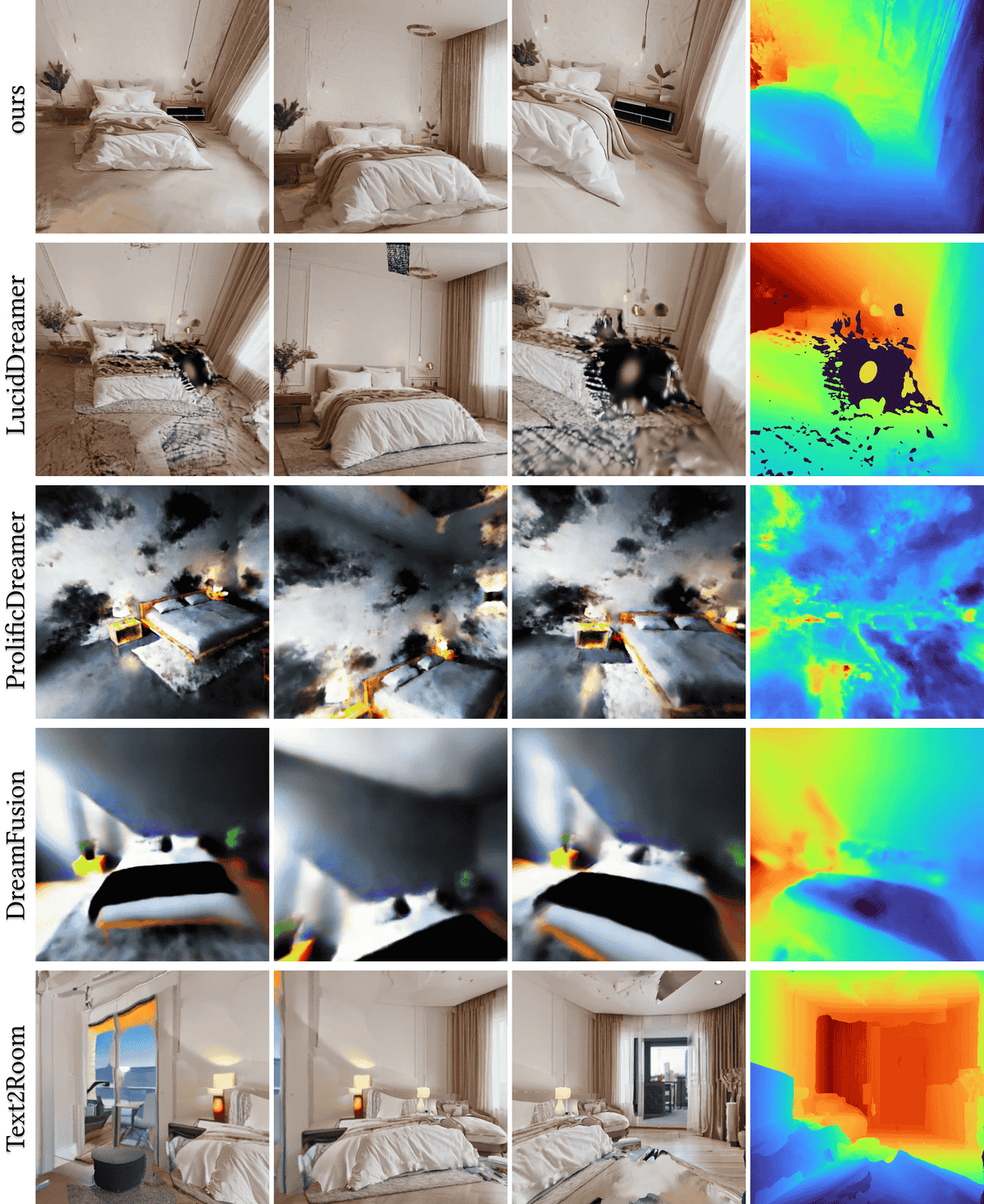}
    \captionsetup{labelformat=empty}
    \caption{\huge Prompt: "A minimalist bedroom, 4K image, high resolution"}
\end{figure*}

\clearpage
\thispagestyle{empty}
\begin{figure*}
    \centering
    \includegraphics[height=1\textheight,keepaspectratio]{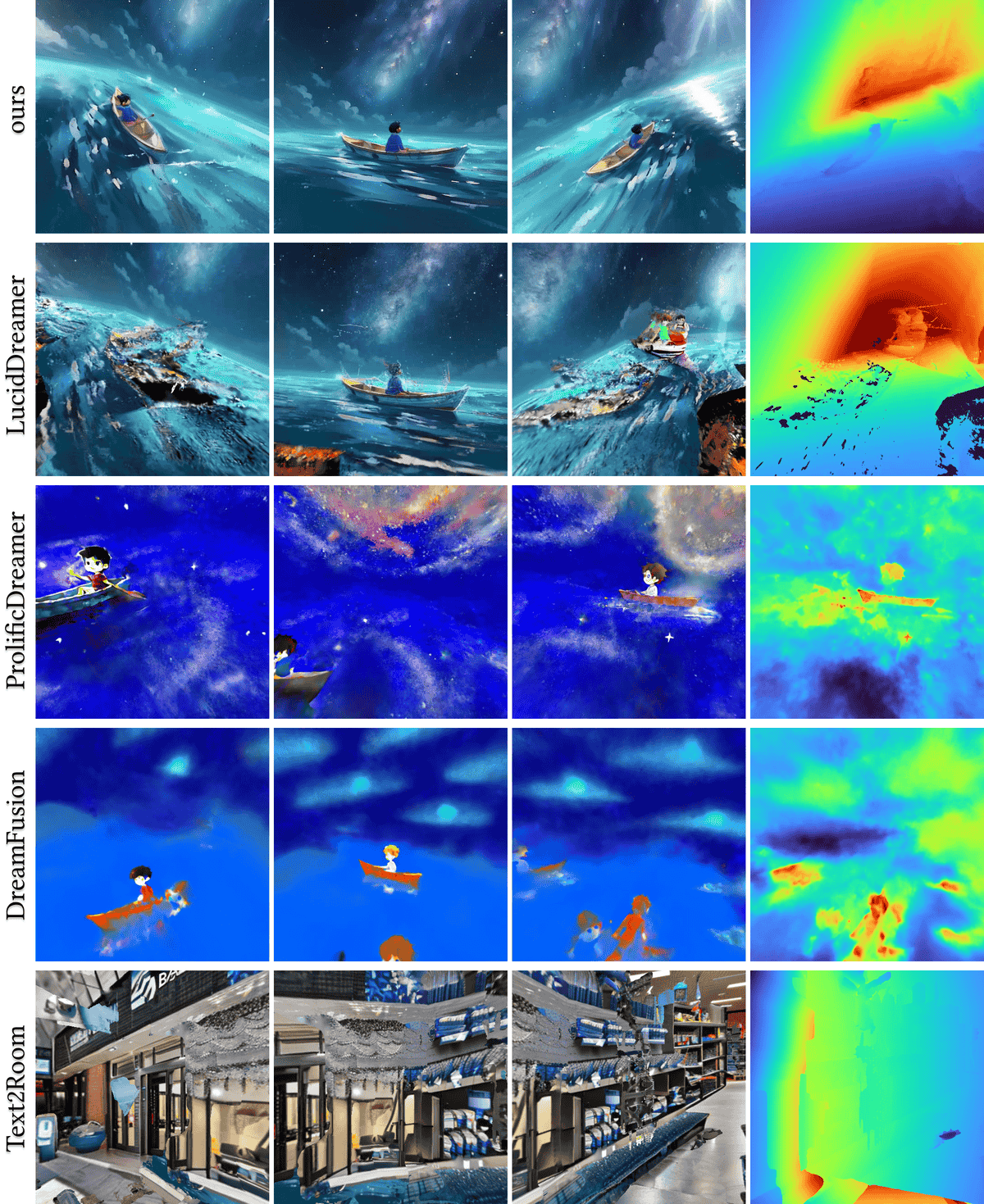}
    \captionsetup{labelformat=empty}
    \caption{\huge Prompt: "A boy sitting in a boat in the middle of the ocean, under the milkyway, anime style"}
\end{figure*}

\clearpage
\thispagestyle{empty}
\begin{figure*}
    \centering
    \includegraphics[height=1\textheight,keepaspectratio]{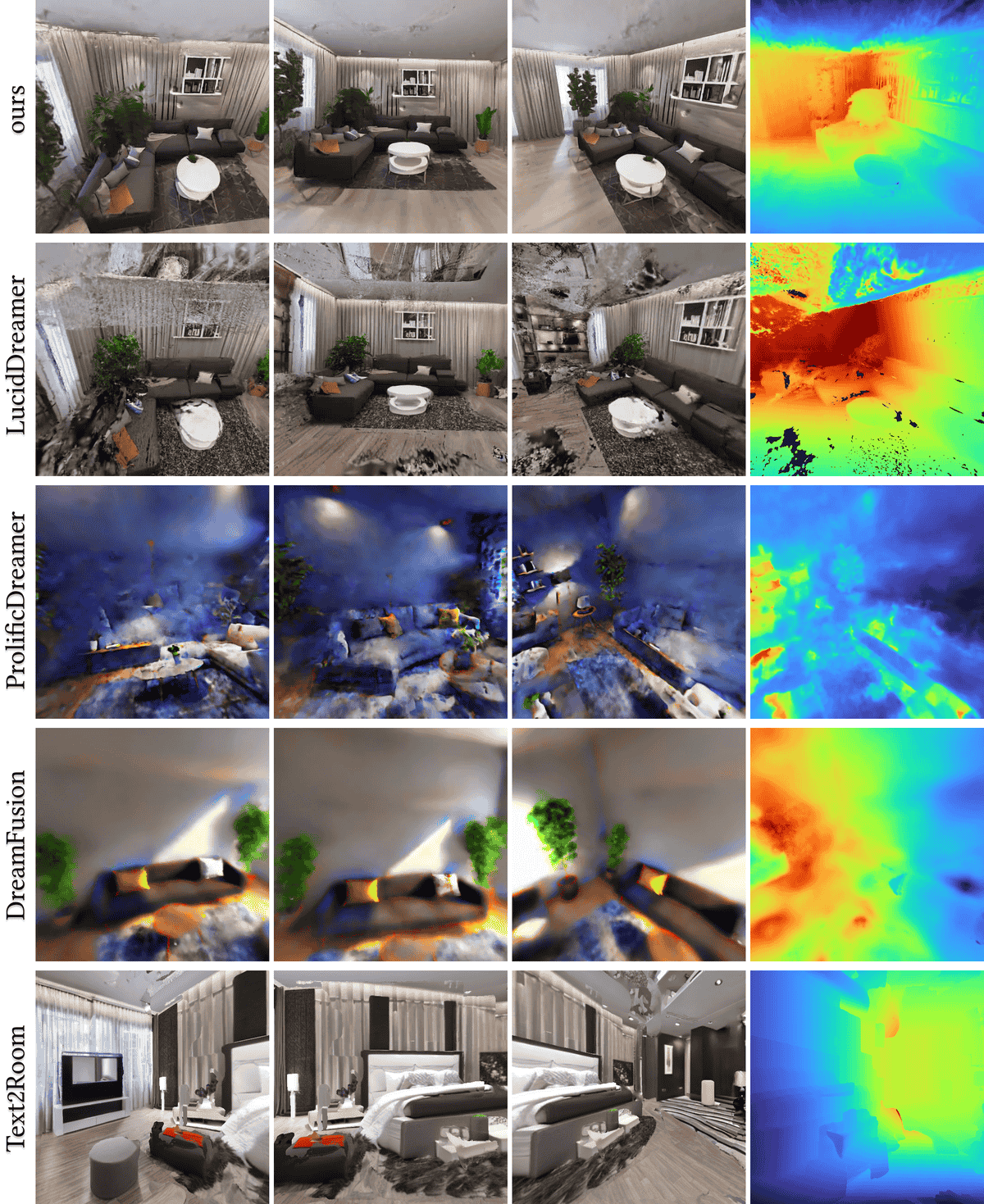}
    \captionsetup{labelformat=empty}
    \caption{\huge Prompt: "a living room, high quality, 8K image, photorealistic"}
\end{figure*}

\clearpage
\thispagestyle{empty}
\begin{figure*}
    \centering
    \includegraphics[height=1\textheight,keepaspectratio]{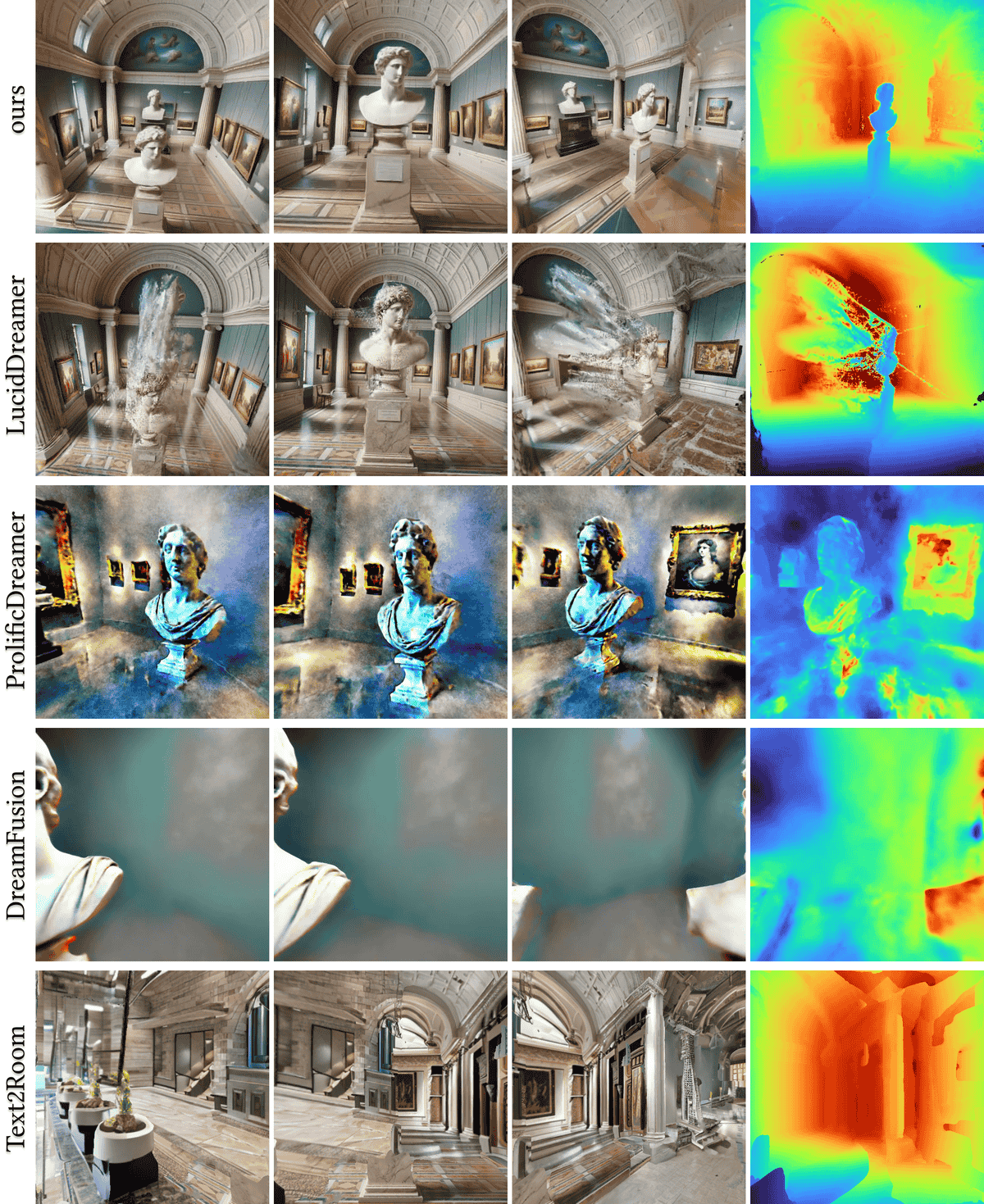}
    \captionsetup{labelformat=empty}
    \caption{\huge Prompt: "A marble bust in a museum with pale teal walls, framed paintings, marble patterned floor, 4k image"}
\end{figure*}

\clearpage
\thispagestyle{empty}
\begin{figure*}
    \centering
    \includegraphics[height=1\textheight,keepaspectratio]{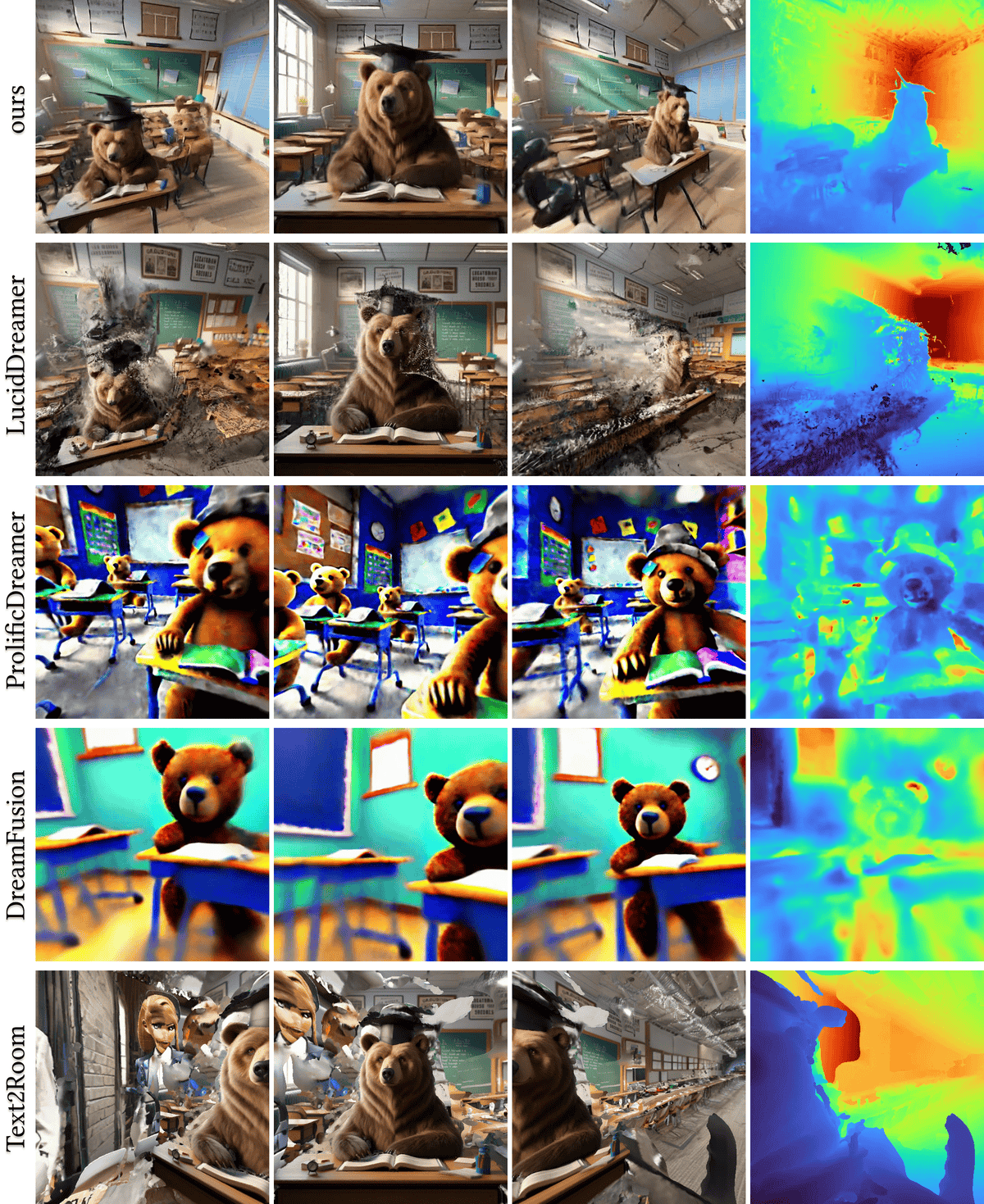}
    \captionsetup{labelformat=empty}
    \caption{\huge Prompt: "A bear sitting in a classroom with a hat on, realistic, 4k image, high detail"}
\end{figure*}

\clearpage
\thispagestyle{empty}
\begin{figure*}
    \centering
    \includegraphics[height=1\textheight,keepaspectratio]{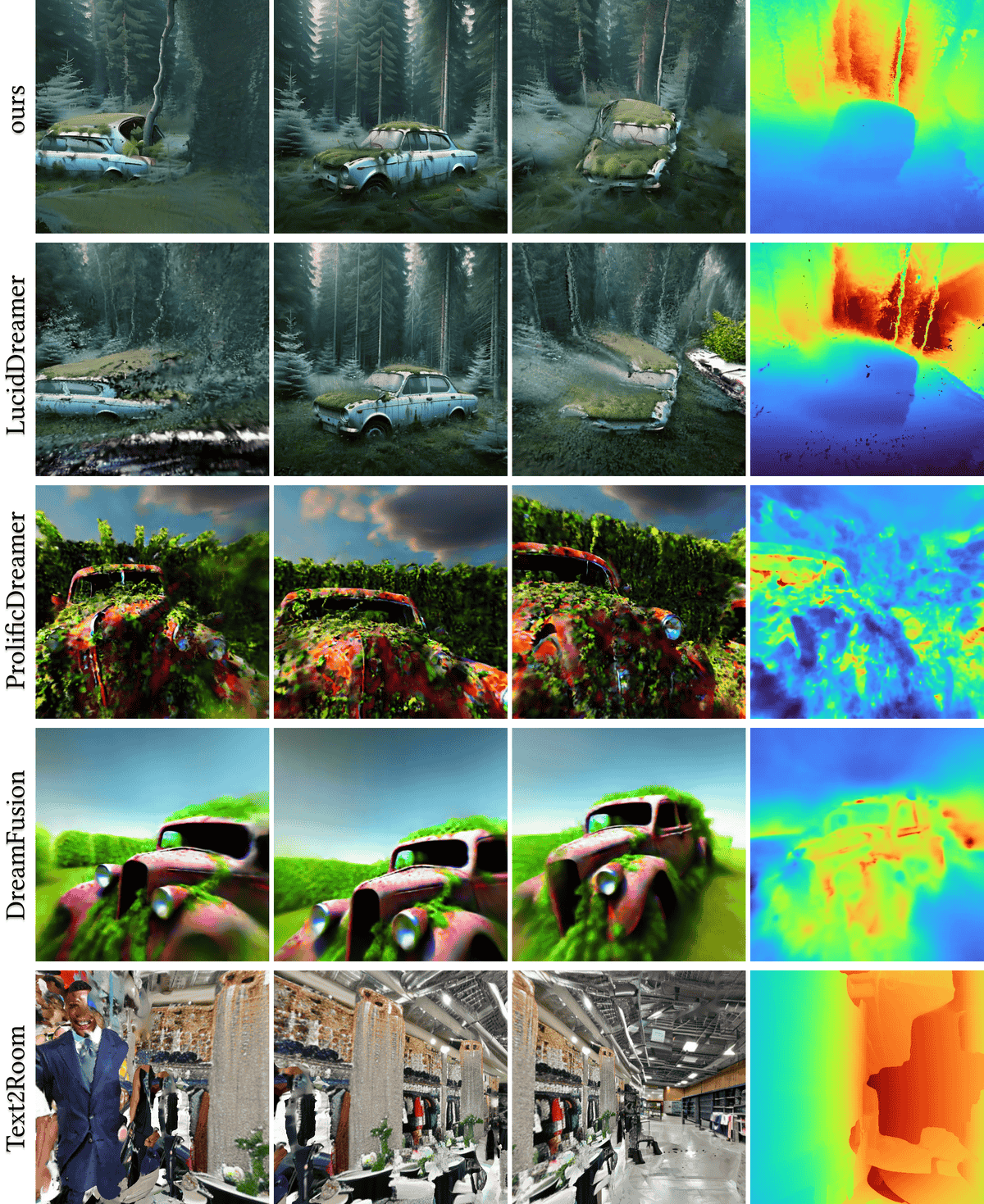}
    \captionsetup{labelformat=empty}
    \caption{\huge Prompt: "An old car overgrown by vines and weeds, high quality image, photorealistic, 4k image"}
\end{figure*}

\clearpage
\thispagestyle{empty}
\begin{figure*}
    \centering
    \includegraphics[height=1\textheight,keepaspectratio]{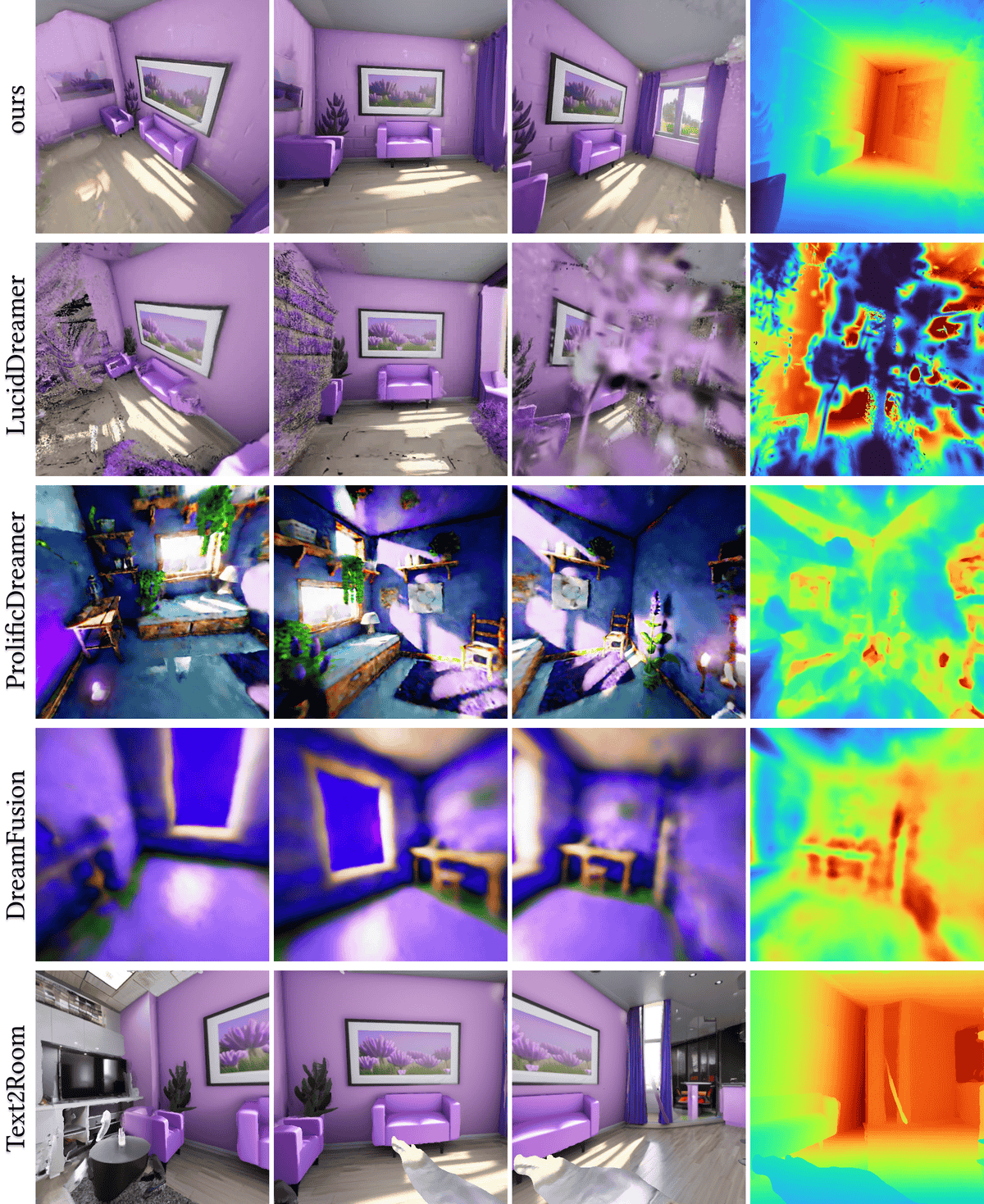}
    \captionsetup{labelformat=empty}
    \caption{\huge Prompt: "Small lavender room, soft lighting, unreal engine render, voxels."}
\end{figure*}

\clearpage
\thispagestyle{empty}
\begin{figure*}
    \centering
    \includegraphics[height=1\textheight,keepaspectratio]{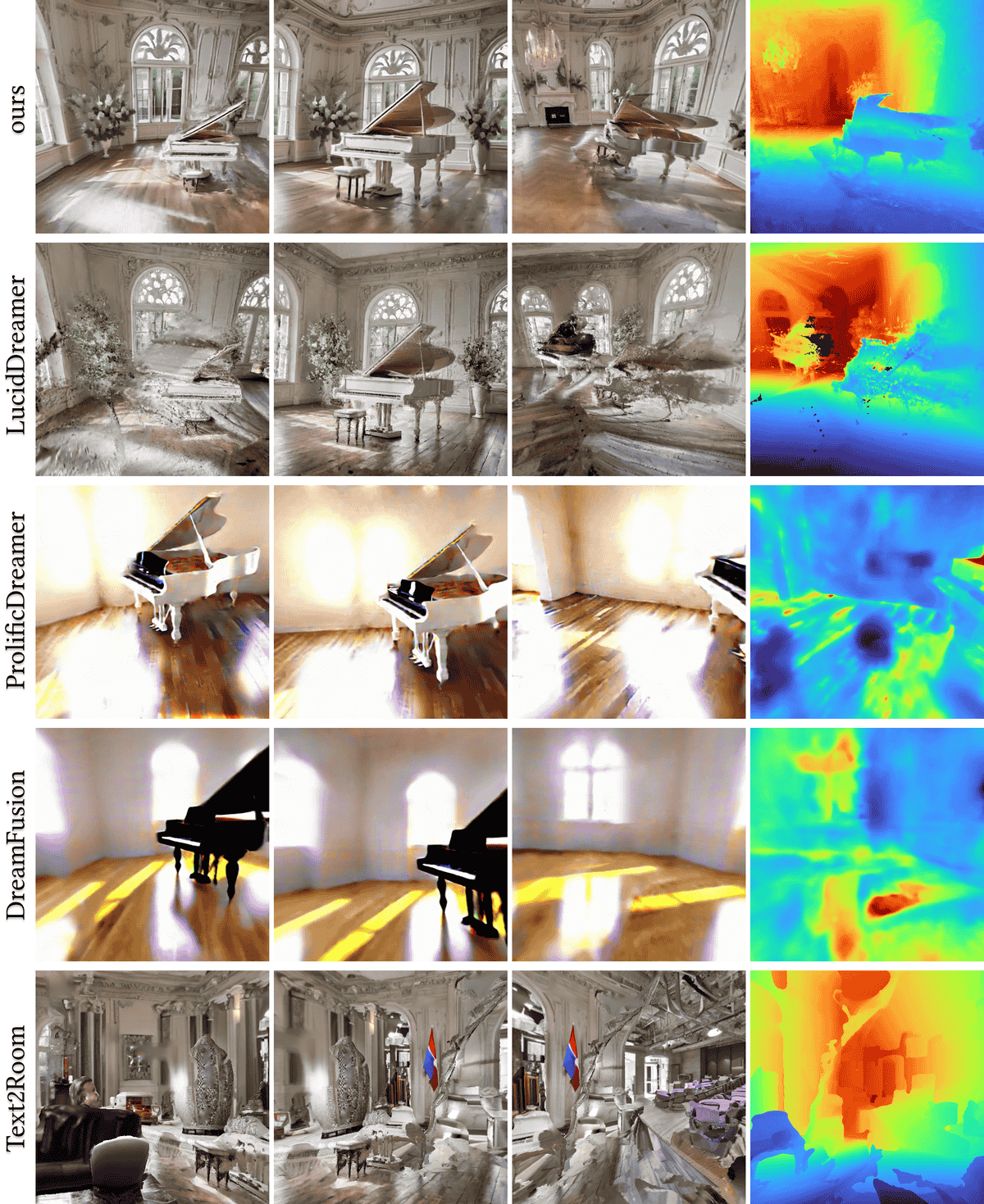}
    \captionsetup{labelformat=empty}
    \caption{\huge Prompt: "White grand piano on wooden floors in an empty hall, 4k image"}
\end{figure*}

\clearpage
\thispagestyle{empty}
\begin{figure*}
    \centering
    \includegraphics[height=1\textheight,keepaspectratio]{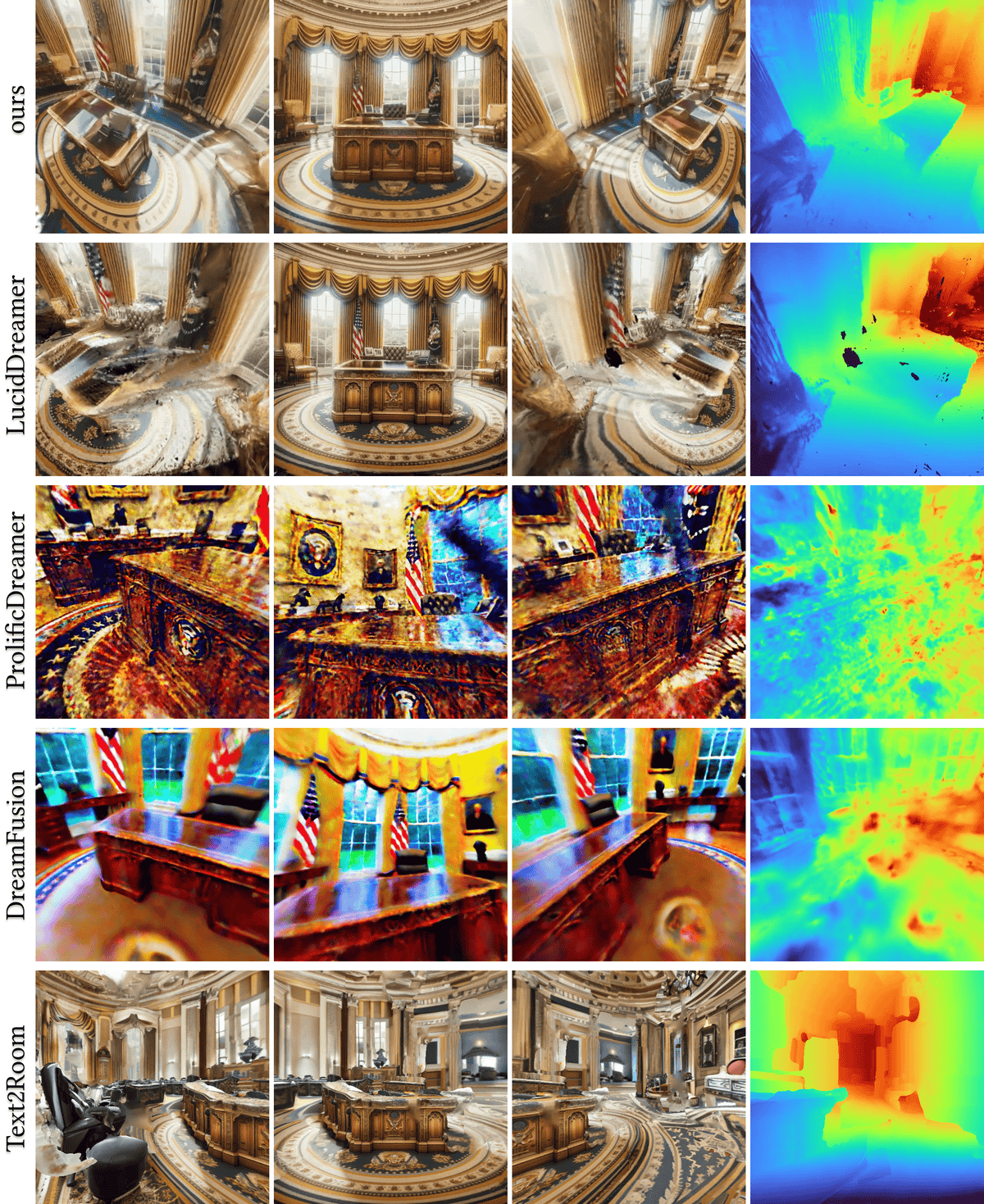}
    \captionsetup{labelformat=empty}
    \caption{\huge Prompt: "A highly detailed image of the resolute desk in the oval office, 4k image"}
\end{figure*}

\clearpage
\thispagestyle{empty}
\begin{figure*}
    \centering
    \includegraphics[height=1\textheight,keepaspectratio]{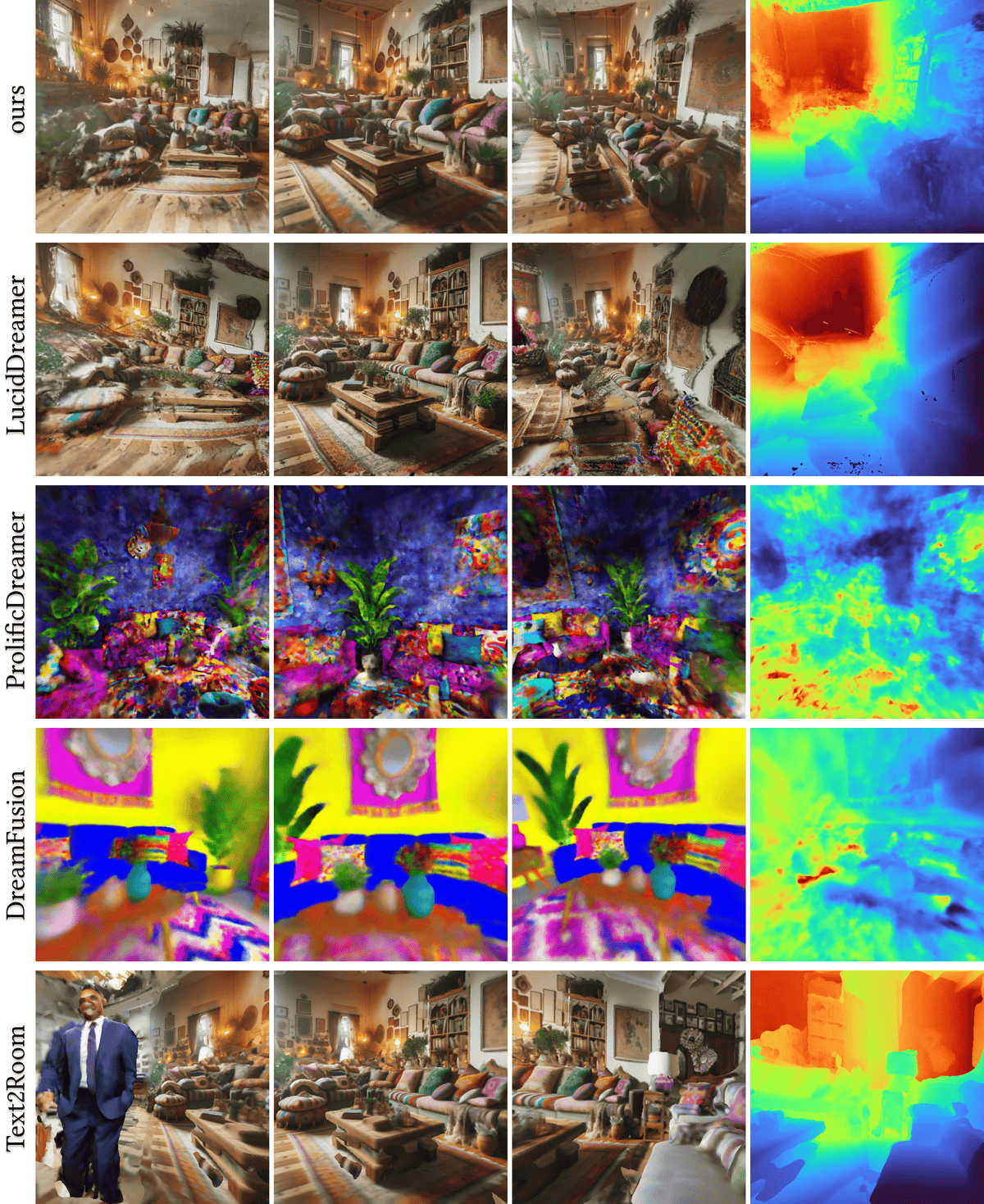}
    \captionsetup{labelformat=empty}
    \caption{\huge Prompt: "A bohemian living room, colorful textiles, vibrant, eclectic, 4k image, photorealistic"}
\end{figure*}

\clearpage
\thispagestyle{empty}
\begin{figure*}
    \centering
    \includegraphics[height=1\textheight,keepaspectratio]{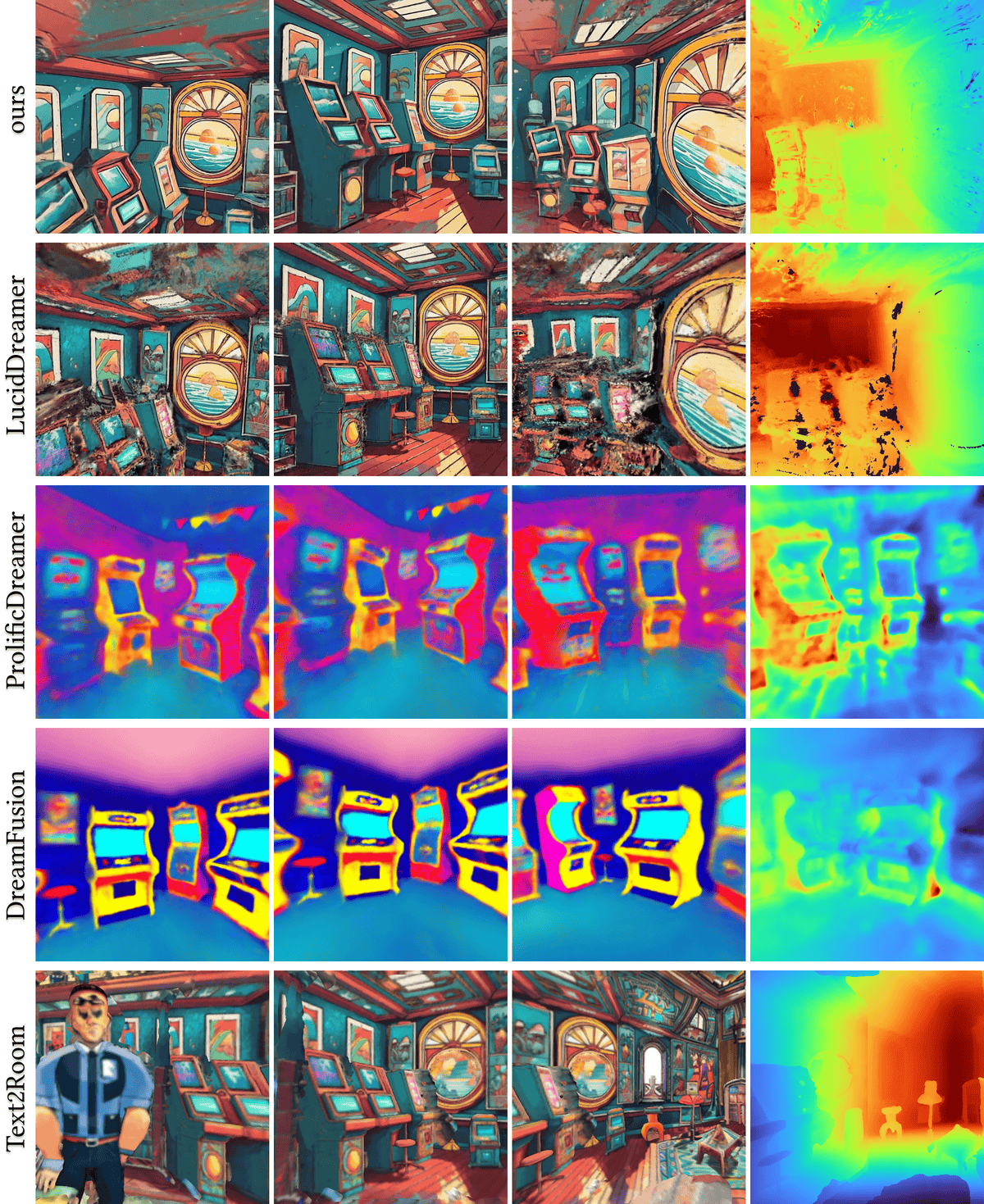}
    \captionsetup{labelformat=empty}
    \caption{\huge Prompt: "Retro arcade room with posters on the walls, retro art style, illustration"}
\end{figure*}

\clearpage
\thispagestyle{empty}
\begin{figure*}
    \centering
    \includegraphics[height=1\textheight,keepaspectratio]{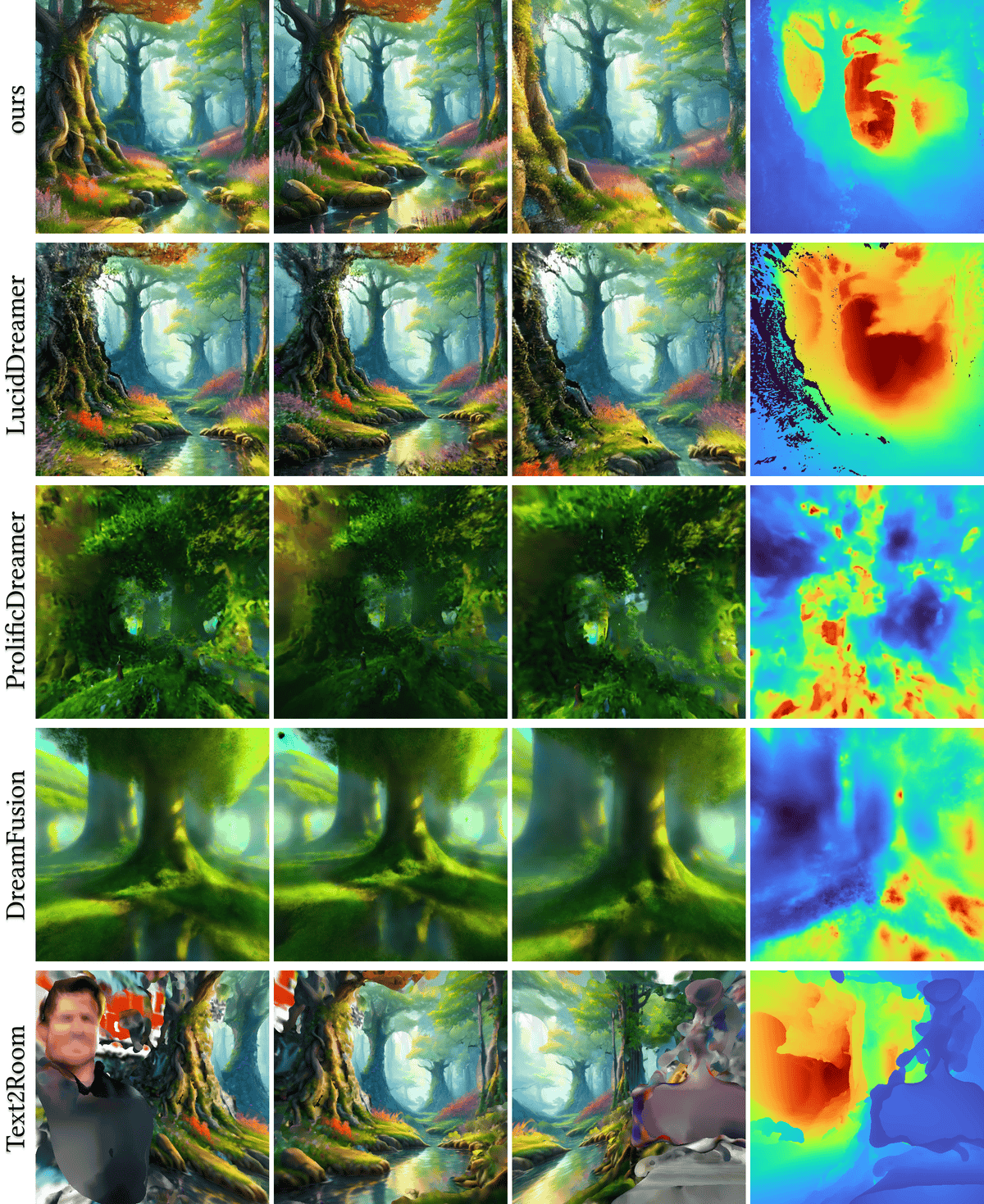}
    \captionsetup{labelformat=empty}
    \caption{\huge Prompt: "A thick elven forest, fantasy art, landscape, picturesque, 4k image"}
\end{figure*}

\clearpage
\thispagestyle{empty}
\begin{figure*}
    \centering
    \includegraphics[height=1\textheight,keepaspectratio]{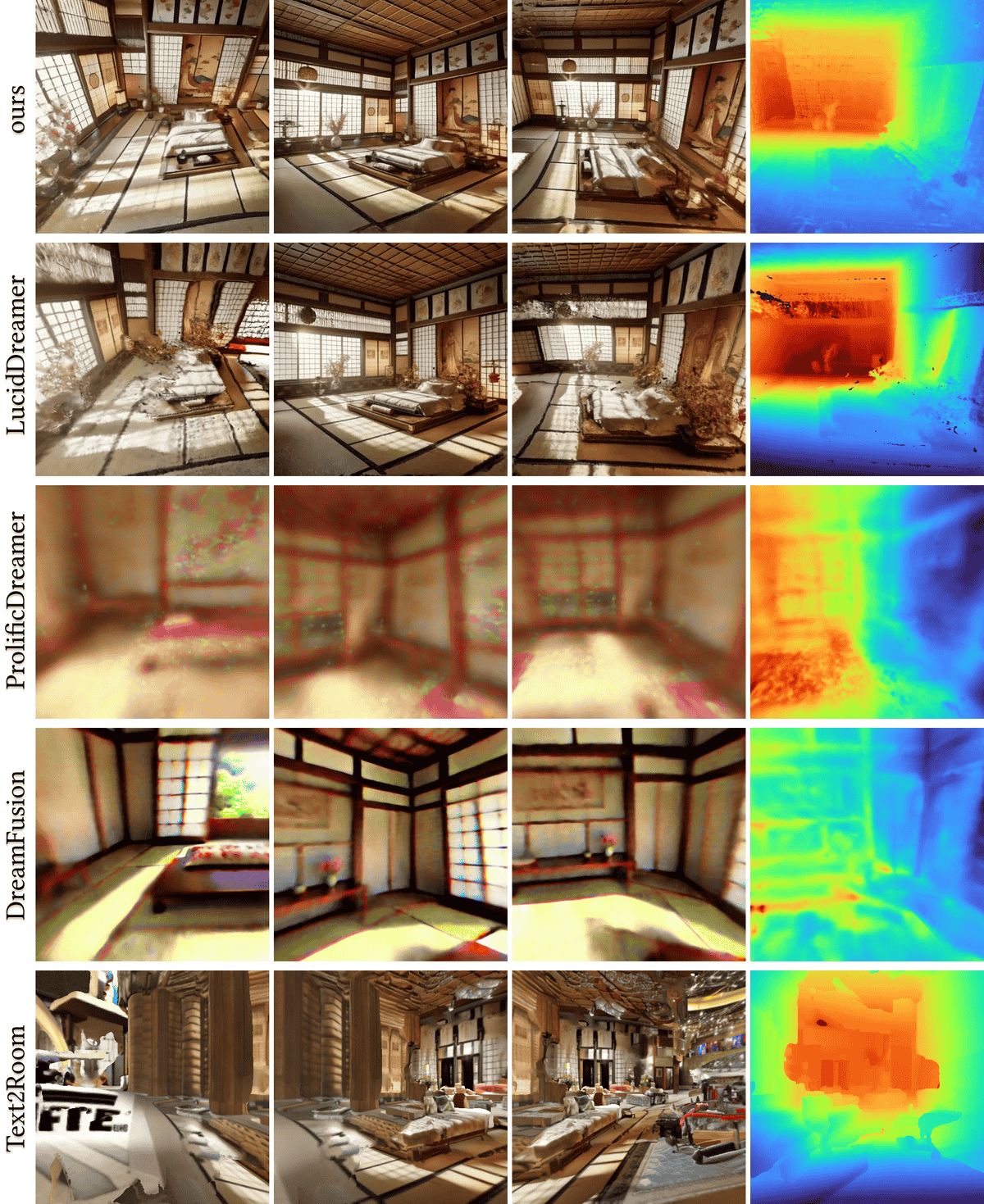}
    \captionsetup{labelformat=empty}
    \caption{\huge Prompt: "A sunny royal traditional Japanese bedroom, 4k image, ornate, high detail"}
\end{figure*}

\clearpage
\thispagestyle{empty}
\begin{figure*}
    \centering
    \includegraphics[height=1\textheight,keepaspectratio]{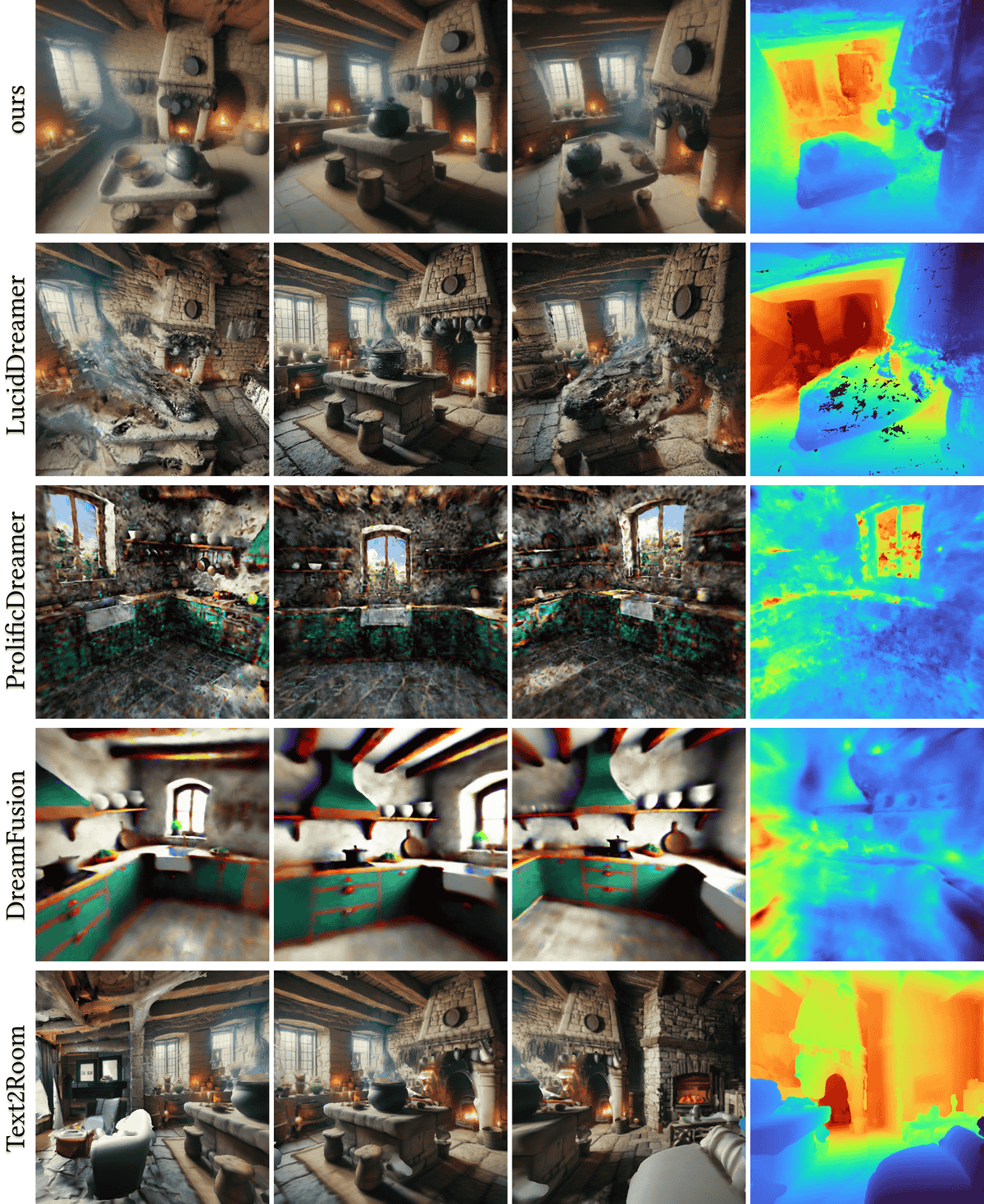}
    \captionsetup{labelformat=empty}
    \caption{\huge Prompt: "An old charming stone kitchen, 4k image, photorealistic, high detail"}
\end{figure*}

\clearpage
\thispagestyle{empty}
\begin{figure*}
    \centering
    \includegraphics[height=1\textheight,keepaspectratio]{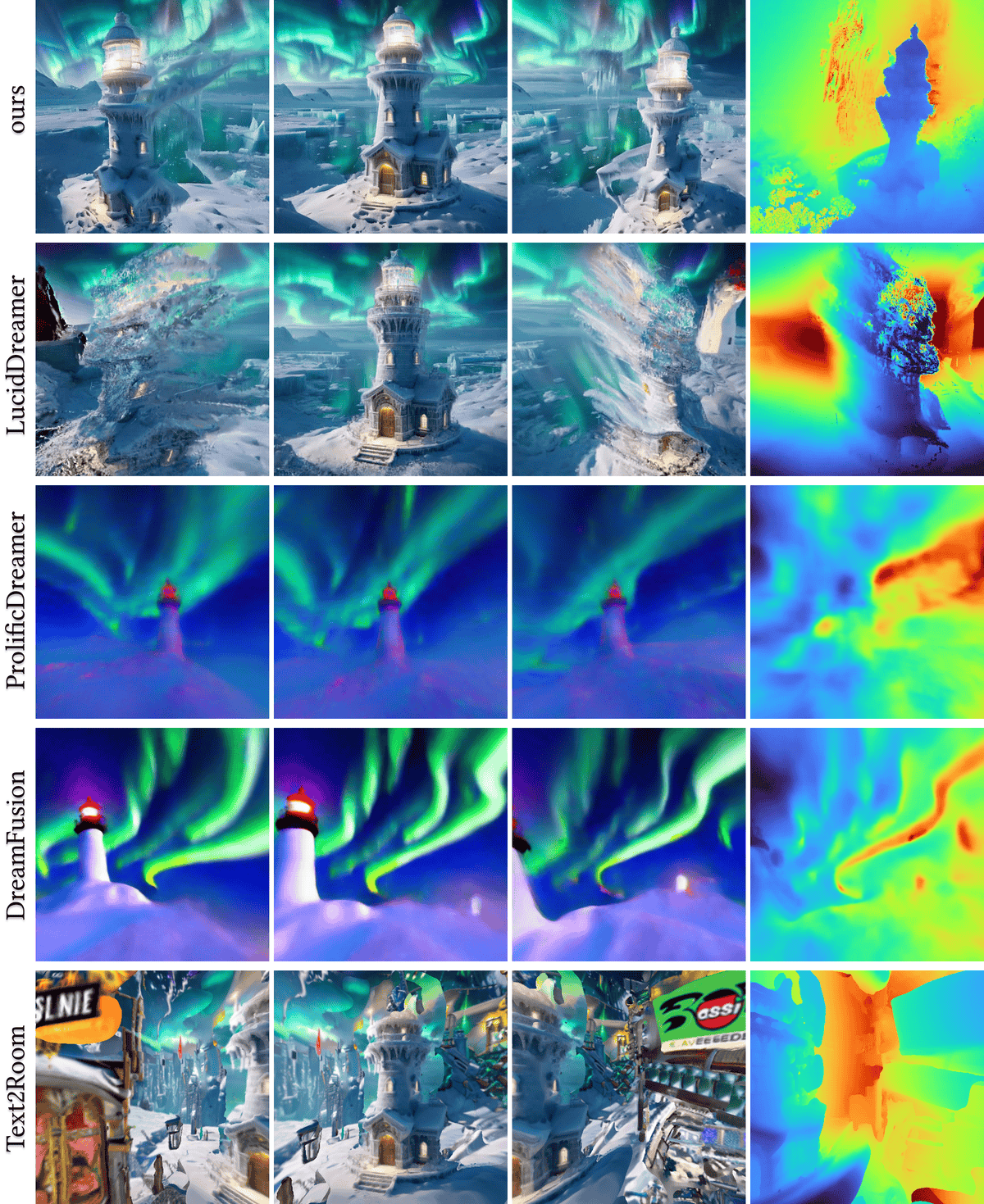}
    \captionsetup{labelformat=empty}
    \caption{\huge Prompt: "Fantasy lighthouse in the Arctic, surrounded by a world of ice and snow, shining with a mystical light under the aurora borealis, 4k, sharp"}
\end{figure*}

\clearpage
\thispagestyle{empty}
\begin{figure*}
    \centering
    \includegraphics[height=1\textheight,keepaspectratio]{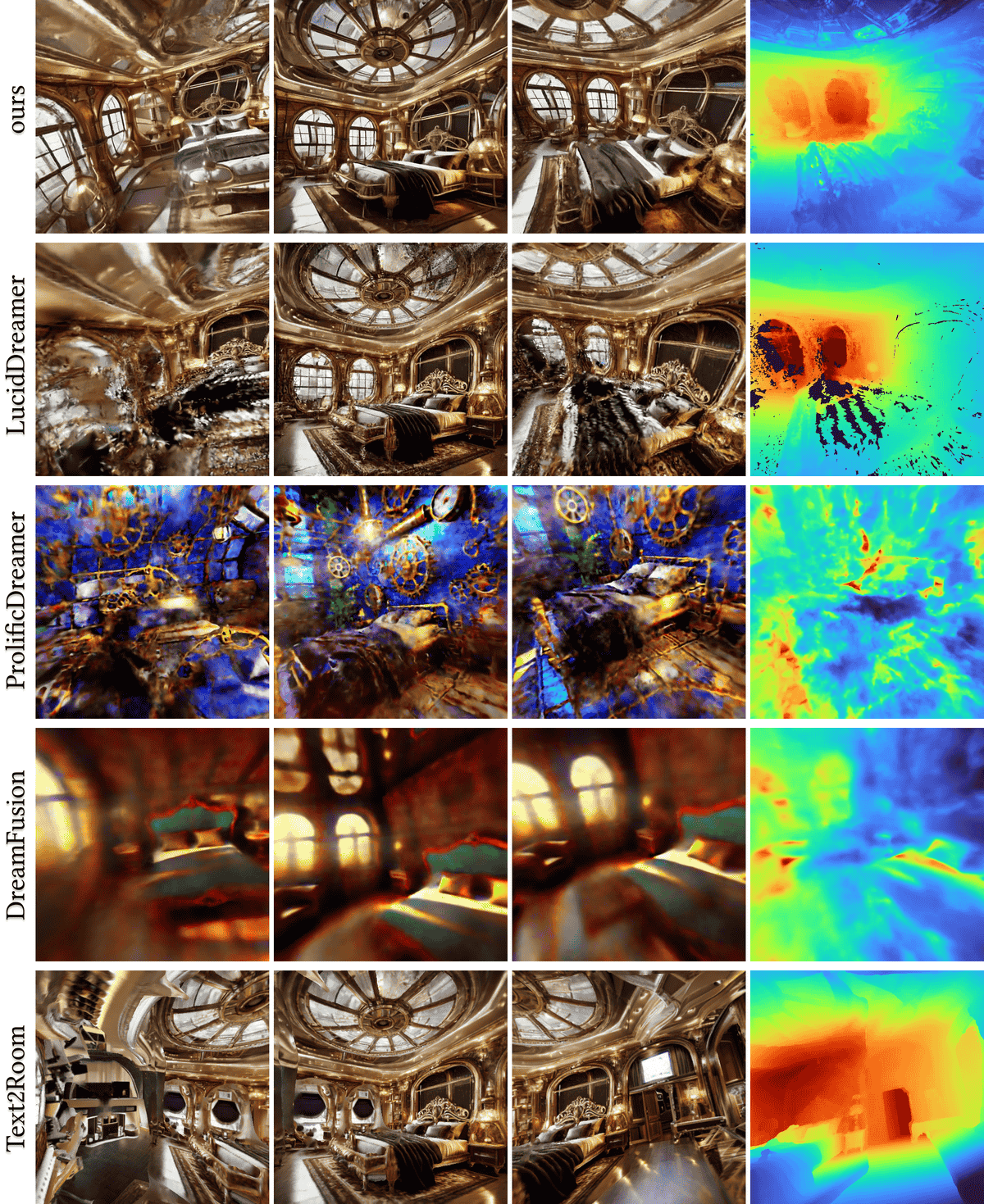}
    \captionsetup{labelformat=empty}
    \caption{\huge Prompt: "A steampunk bedroom with glass ceilings, photorealistic, 4k image, bright lighting"}
\end{figure*}

\clearpage
\thispagestyle{empty}
\begin{figure*}
    \centering
    \includegraphics[height=1\textheight,keepaspectratio]{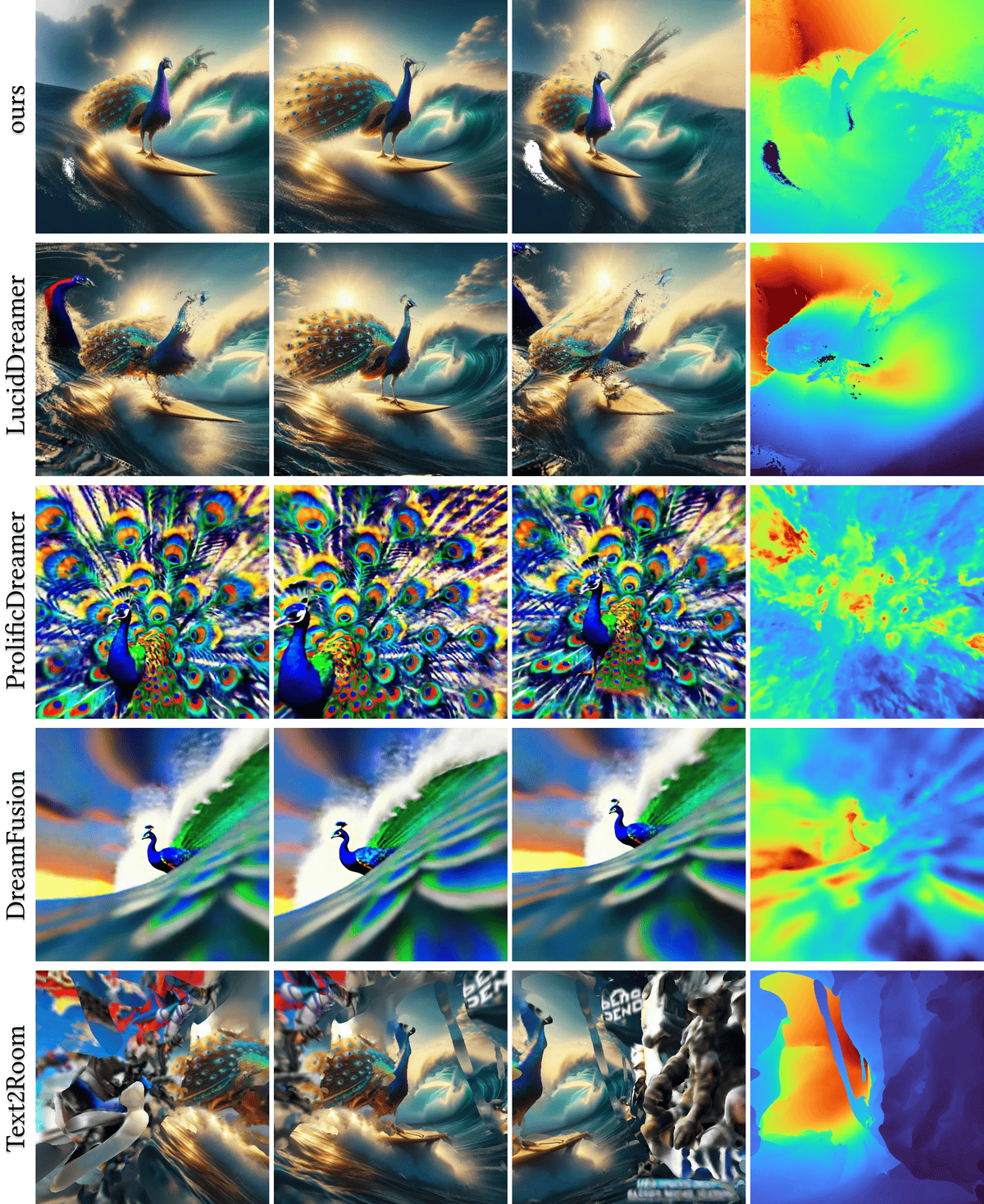}
    \captionsetup{labelformat=empty}
    \caption{\huge Prompt: "A majestic peacock, surfing a tall wave, photorealistic, detailed image, 4k image"}
\end{figure*}

\clearpage
\thispagestyle{empty}
\begin{figure*}
    \centering
    \includegraphics[height=1\textheight,keepaspectratio]{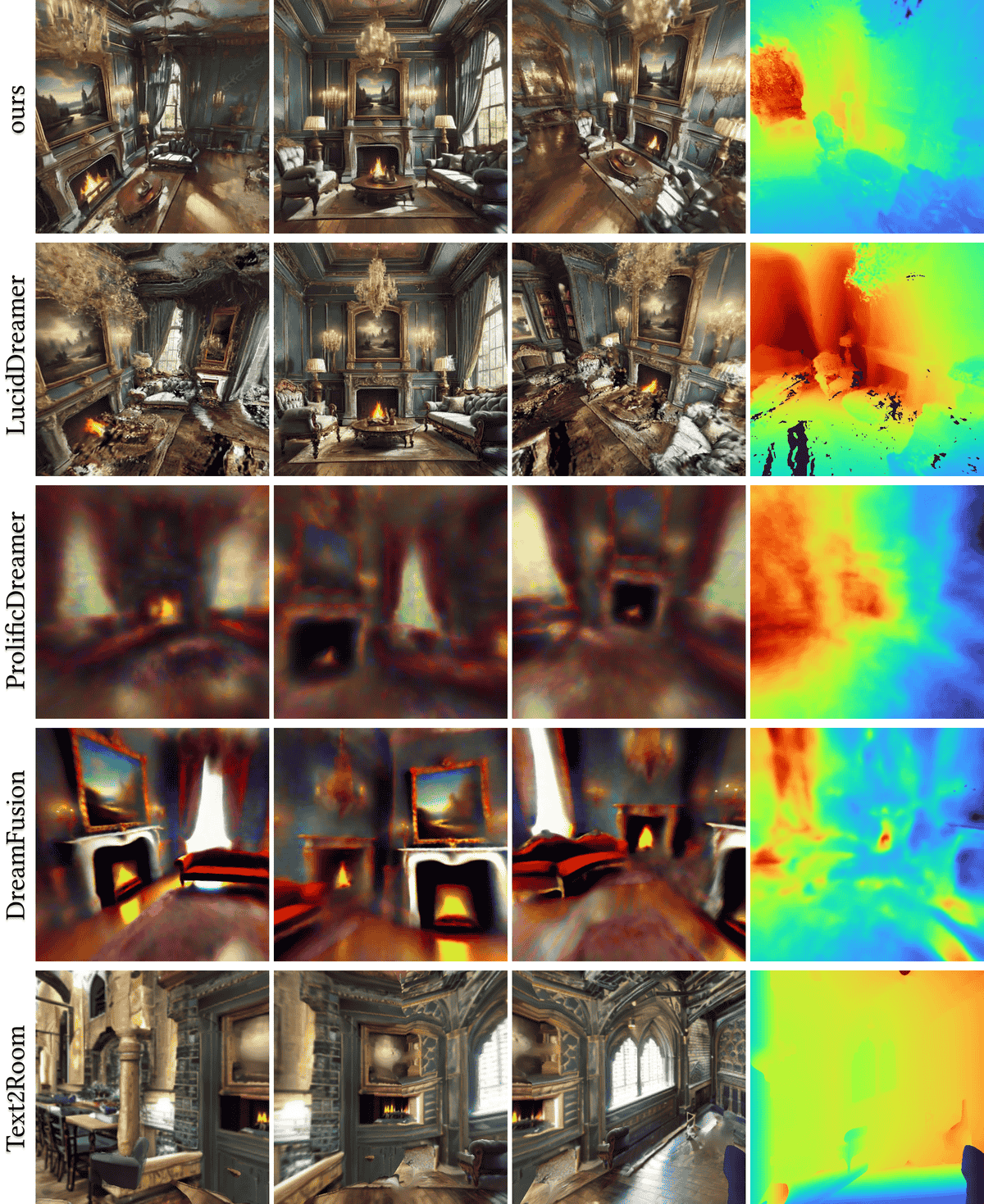}
    \captionsetup{labelformat=empty}
    \caption{\huge Prompt: "A victorian living room with a grand fireplace and a long sofa, painting over the fireplace, mysterious vibe, giant windows, 4k image, photorealistic"}
\end{figure*}
\thispagestyle{empty}
\clearpage

%% file: main.bbl
\begin{thebibliography}{72}
\providecommand{\natexlab}[1]{#1}
\providecommand{\url}[1]{\texttt{#1}}
\expandafter\ifx\csname urlstyle\endcsname\relax
  \providecommand{\doi}[1]{doi: #1}\else
  \providecommand{\doi}{doi: \begingroup \urlstyle{rm}\Url}\fi

\bibitem[Arjovsky et~al.(2017)Arjovsky, Chintala, and Bottou]{wgan}
Martin Arjovsky, Soumith Chintala, and L{\'e}on Bottou.
\newblock Wasserstein generative adversarial networks.
\newblock In \emph{International Conference on Machine Learning (ICML)}, pages
  214--223, 2017.

\bibitem[Armandpour et~al.(2023)Armandpour, Zheng, Sadeghian, Sadeghian, and
  Zhou]{armandpour2023re}
Mohammadreza Armandpour, Huangjie Zheng, Ali Sadeghian, Amir Sadeghian, and
  Mingyuan Zhou.
\newblock Re-imagine the negative prompt algorithm: Transform 2d diffusion into
  3d, alleviate janus problem and beyond.
\newblock \emph{arXiv preprint arXiv:2304.04968}, 2023.

\bibitem[Bae et~al.(2022)Bae, Budvytis, and Cipolla]{irondepth}
Gwangbin Bae, Ignas Budvytis, and Roberto Cipolla.
\newblock Irondepth: Iterative refinement of single-view depth using surface
  normal and its uncertainty.
\newblock In \emph{British Machine Vision Conference (BMVC)}, 2022.

\bibitem[Betker et~al.(2023)Betker, Goh, Jing, Brooks, Wang, Li, Ouyang,
  Zhuang, Lee, and Guo]{dalle3}
James Betker, Gabriel Goh, Li Jing, Tim Brooks, Jianfeng Wang, Linjie Li, Long
  Ouyang, Juntang Zhuang, Joyce Lee, and Yufei et~al. Guo.
\newblock Improving image generation with better captions.
\newblock page~8, 2023.

\bibitem[Bhat et~al.(2023)Bhat, Birkl, Wofk, Wonka, and Müller]{zoedepth}
Shariq~Farooq Bhat, Reiner Birkl, Diana Wofk, Peter Wonka, and Matthias
  Müller.
\newblock Zoedepth: Zero-shot transfer by combining relative and metric depth.
\newblock \emph{arXiv preprint arXiv:2302.12288}, 2023.

\bibitem[Brooks et~al.(2023)Brooks, Holynski, and Efros]{pix2pix}
Tim Brooks, Aleksander Holynski, and Alexei~A Efros.
\newblock Instructpix2pix: Learning to follow image editing instructions.
\newblock In \emph{Conference on Computer Vision and Pattern Recognition
  (CVPR)}, pages 18392--18402, 2023.

\bibitem[Bylinskii et~al.(2022)Bylinskii, Herman, Hertzmann, Hutka, and
  Zhang]{Bylinskii2022TowardsBU}
Zoya Bylinskii, Laura~Mariah Herman, Aaron Hertzmann, Stefanie Hutka, and Yile
  Zhang.
\newblock Towards better user studies in computer graphics and vision.
\newblock \emph{Found. Trends Comput. Graph. Vis.}, 15:\penalty0 201--252,
  2022.

\bibitem[Chan et~al.(2023)Chan, Nagano, Chan, Bergman, Park, Levy, Aittala,
  De~Mello, Karras, and Wetzstein]{gennvs}
Eric~R Chan, Koki Nagano, Matthew~A Chan, Alexander~W Bergman, Jeong~Joon Park,
  Axel Levy, Miika Aittala, Shalini De~Mello, Tero Karras, and Gordon
  Wetzstein.
\newblock Generative novel view synthesis with 3d-aware diffusion models.
\newblock \emph{arXiv preprint arXiv:2304.02602}, 2023.

\bibitem[Chang et~al.(2014)Chang, Savva, and Manning]{retrieval1}
Angel Chang, Manolis Savva, and Christopher~D Manning.
\newblock Learning spatial knowledge for text to 3d scene generation.
\newblock pages 2028--2038, 2014.

\bibitem[Chang et~al.(2015)Chang, Monroe, Savva, Potts, and
  Manning]{retrieval2}
Angel Chang, Will Monroe, Manolis Savva, Christopher Potts, and Christopher~D
  Manning.
\newblock Text to 3d scene generation with rich lexical grounding.
\newblock \emph{arXiv preprint arXiv:1505.06289}, 2015.

\bibitem[Chen et~al.(2019)Chen, Choy, Savva, Chang, Funkhouser, and
  Savarese]{text2shape}
Kevin Chen, Christopher~B Choy, Manolis Savva, Angel~X Chang, Thomas
  Funkhouser, and Silvio Savarese.
\newblock Text2shape: Generating shapes from natural language by learning joint
  embeddings.
\newblock In \emph{Asian Conference on Computer Vision (ACCV)}, pages 100--116.
  Springer, 2019.

\bibitem[Chen et~al.(2023{\natexlab{a}})Chen, Chen, Jiao, and Jia]{fantasia}
Rui Chen, Yongwei Chen, Ningxin Jiao, and Kui Jia.
\newblock Fantasia3d: Disentangling geometry and appearance for high-quality
  text-to-3d content creation.
\newblock In \emph{International Conference on Computer Vision (ICCV)},
  2023{\natexlab{a}}.

\bibitem[Chen et~al.(2023{\natexlab{b}})Chen, Wang, and Liu]{gsgen}
Zilong Chen, Feng Wang, and Huaping Liu.
\newblock Text-to-3d using gaussian splatting.
\newblock \emph{arXiv preprint arXiv:2309.16585}, 2023{\natexlab{b}}.

\bibitem[Chung et~al.(2023)Chung, Lee, Nam, Lee, and Lee]{luciddreamer}
Jaeyoung Chung, Suyoung Lee, Hyeongjin Nam, Jaerin Lee, and Kyoung~Mu Lee.
\newblock Luciddreamer: Domain-free generation of 3d gaussian splatting scenes.
\newblock \emph{arXiv preprint arXiv:2311.13384}, 2023.

\bibitem[Coyne and Sproat(2001)]{wrodseye}
Bob Coyne and Richard Sproat.
\newblock Wordseye: An automatic text-to-scene conversion system.
\newblock pages 487--496, 2001.

\bibitem[Deitke et~al.(2022)Deitke, Schwenk, Salvador, Weihs, Michel,
  VanderBilt, Schmidt, Ehsani, Kembhavi, and Farhadi]{Deitke2022ObjaverseAU}
Matt Deitke, Dustin Schwenk, Jordi Salvador, Luca Weihs, Oscar Michel, Eli
  VanderBilt, Ludwig Schmidt, Kiana Ehsani, Aniruddha Kembhavi, and Ali
  Farhadi.
\newblock Objaverse: A universe of annotated 3d objects.
\newblock In \emph{Conference on Computer Vision and Pattern Recognition
  (CVPR)}, pages 13142--13153, 2022.

\bibitem[Deng et~al.(2023)Deng, Jiang, Qi, Yan, Zhou, Guibas, Anguelov,
  et~al.]{nerdi}
Congyue Deng, Chiyu Jiang, Charles~R Qi, Xinchen Yan, Yin Zhou, Leonidas
  Guibas, Dragomir Anguelov, et~al.
\newblock Nerdi: Single-view nerf synthesis with language-guided diffusion as
  general image priors.
\newblock In \emph{Conference on Computer Vision and Pattern Recognition
  (CVPR)}, pages 20637--20647, 2023.

\bibitem[Fridman et~al.(2023)Fridman, Abecasis, Kasten, and Dekel]{SceneScape}
Rafail Fridman, Amit Abecasis, Yoni Kasten, and Tali Dekel.
\newblock Scenescape: Text-driven consistent scene generation.
\newblock \emph{arXiv preprint arXiv:2302.01133}, 2023.

\bibitem[Fridovich-Keil et~al.(2022)Fridovich-Keil, Yu, Tancik, Chen, Recht,
  and Kanazawa]{plenoxels}
Sara Fridovich-Keil, Alex Yu, Matthew Tancik, Qinhong Chen, Benjamin Recht, and
  Angjoo Kanazawa.
\newblock Plenoxels: Radiance fields without neural networks.
\newblock In \emph{Conference on Computer Vision and Pattern Recognition
  (CVPR)}, pages 5501--5510, 2022.

\bibitem[Gao et~al.(2024)Gao, Holynski, Henzler, Brussee, Martin-Brualla,
  Srinivasan, Barron, and Poole]{cat3d}
Ruiqi Gao, Aleksander Holynski, Philipp Henzler, Arthur Brussee, Ricardo
  Martin-Brualla, Pratul~P. Srinivasan, Jonathan~T. Barron, and Ben Poole.
\newblock Cat3d: Create anything in 3d with multi-view diffusion models.
\newblock \emph{arXiv}, 2024.

\bibitem[Gu et~al.(2023)Gu, Trevithick, Lin, Susskind, Theobalt, Liu, and
  Ramamoorthi]{nerfdiff}
Jiatao Gu, Alex Trevithick, Kai-En Lin, Joshua~M Susskind, Christian Theobalt,
  Lingjie Liu, and Ravi Ramamoorthi.
\newblock Nerfdiff: Single-image view synthesis with nerf-guided distillation
  from 3d-aware diffusion.
\newblock In \emph{International Conference on Machine Learning (ICML)}, pages
  11808--11826, 2023.

\bibitem[Guo et~al.(2023)Guo, Liu, Shao, Laforte, Voleti, Luo, Chen, Zou, Wang,
  Cao, and Zhang]{threestudio2023}
Yuan-Chen Guo, Ying-Tian Liu, Ruizhi Shao, Christian Laforte, Vikram Voleti,
  Guan Luo, Chia-Hao Chen, Zi-Xin Zou, Chen Wang, Yan-Pei Cao, and Song-Hai
  Zhang.
\newblock threestudio: A unified framework for 3d content generation.
\newblock \url{https://github.com/threestudio-project/threestudio}, 2023.

\bibitem[Hertz et~al.(2022)Hertz, Mokady, Tenenbaum, Aberman, Pritch, and
  Cohen-Or]{hertz2022prompt}
Amir Hertz, Ron Mokady, Jay Tenenbaum, Kfir Aberman, Yael Pritch, and Daniel
  Cohen-Or.
\newblock Prompt-to-prompt image editing with cross attention control.
\newblock \emph{arXiv preprint arXiv:2208.01626}, 2022.

\bibitem[Ho and Salimans(2022)]{cfg}
Jonathan Ho and Tim Salimans.
\newblock Classifier-free diffusion guidance.
\newblock \emph{arXiv preprint arXiv:2207.12598}, 2022.

\bibitem[Ho et~al.(2020)Ho, Jain, and Abbeel]{ddpm}
Jonathan Ho, Ajay Jain, and Pieter Abbeel.
\newblock Denoising diffusion probabilistic models.
\newblock \emph{NIPS}, 33:\penalty0 6840--6851, 2020.

\bibitem[H\"ollein et~al.(2023)H\"ollein, Cao, Owens, Johnson, and
  Nie{\ss}ner]{hoellein2023text2room}
Lukas H\"ollein, Ang Cao, Andrew Owens, Justin Johnson, and Matthias
  Nie{\ss}ner.
\newblock Text2room: Extracting textured 3d meshes from 2d text-to-image
  models.
\newblock In \emph{International Conference on Computer Vision (ICCV)}, pages
  7909--7920, 2023.

\bibitem[Jain et~al.(2022)Jain, Mildenhall, Barron, Abbeel, and
  Poole]{dreamfields}
Ajay Jain, Ben Mildenhall, Jonathan~T Barron, Pieter Abbeel, and Ben Poole.
\newblock Zero-shot text-guided object generation with dream fields.
\newblock In \emph{Conference on Computer Vision and Pattern Recognition
  (CVPR)}, pages 867--876, 2022.

\bibitem[Kant et~al.(2023)Kant, Siarohin, Vasilkovsky, Guler, Ren, Tulyakov,
  and Gilitschenski]{invs}
Yash Kant, Aliaksandr Siarohin, Michael Vasilkovsky, Riza~Alp Guler, Jian Ren,
  Sergey Tulyakov, and Igor Gilitschenski.
\newblock invs: Repurposing diffusion inpainters for novel view synthesis.
\newblock In \emph{SIGGRAPH Asia 2023}, pages 1--12, 2023.

\bibitem[Karras et~al.(2022)Karras, Aittala, Aila, and Laine]{edm}
Tero Karras, Miika Aittala, Timo Aila, and Samuli Laine.
\newblock Elucidating the design space of diffusion-based generative models.
\newblock \emph{NIPS}, 35:\penalty0 26565--26577, 2022.

\bibitem[Ke et~al.(2023)Ke, Obukhov, Huang, Metzger, Daudt, and
  Schindler]{ke2023repurposing}
Bingxin Ke, Anton Obukhov, Shengyu Huang, Nando Metzger, Rodrigo~Caye Daudt,
  and Konrad Schindler.
\newblock Repurposing diffusion-based image generators for monocular depth
  estimation.
\newblock \emph{arXiv preprint arXiv:2312.02145}, 2023.

\bibitem[Kerbl et~al.(2023)Kerbl, Kopanas, Leimk{\"u}hler, and Drettakis]{3dgs}
Bernhard Kerbl, Georgios Kopanas, Thomas Leimk{\"u}hler, and George Drettakis.
\newblock 3d gaussian splatting for real-time radiance field rendering.
\newblock \emph{ACM TOG}, 42\penalty0 (4), 2023.

\bibitem[Lei et~al.(2023)Lei, Tang, and Jia]{rgbd2}
Jiabao Lei, Jiapeng Tang, and Kui Jia.
\newblock Rgbd2: Generative scene synthesis via incremental view inpainting
  using rgbd diffusion models.
\newblock In \emph{Conference on Computer Vision and Pattern Recognition
  (CVPR)}, pages 8422--8434, 2023.

\bibitem[Liang et~al.(2023)Liang, Yang, Lin, Li, Xu, and
  Chen]{EnVision2023luciddreamer}
Yixun Liang, Xin Yang, Jiantao Lin, Haodong Li, Xiaogang Xu, and Yingcong Chen.
\newblock Luciddreamer: Towards high-fidelity text-to-3d generation via
  interval score matching.
\newblock \emph{arXiv preprint arXiv:2311.11284}, 2023.

\bibitem[Lin et~al.(2023{\natexlab{a}})Lin, Gao, Tang, Takikawa, Zeng, Huang,
  Kreis, Fidler, Liu, and Lin]{lin2023magic3d}
Chen-Hsuan Lin, Jun Gao, Luming Tang, Towaki Takikawa, Xiaohui Zeng, Xun Huang,
  Karsten Kreis, Sanja Fidler, Ming-Yu Liu, and Tsung-Yi Lin.
\newblock Magic3d: High-resolution text-to-3d content creation.
\newblock In \emph{Conference on Computer Vision and Pattern Recognition
  (CVPR)}, 2023{\natexlab{a}}.

\bibitem[Lin et~al.(2023{\natexlab{b}})Lin, Gao, Tang, Takikawa, Zeng, Huang,
  Kreis, Fidler, Liu, and Lin]{magic3d}
Chen-Hsuan Lin, Jun Gao, Luming Tang, Towaki Takikawa, Xiaohui Zeng, Xun Huang,
  Karsten Kreis, Sanja Fidler, Ming-Yu Liu, and Tsung-Yi Lin.
\newblock Magic3d: High-resolution text-to-3d content creation.
\newblock In \emph{Conference on Computer Vision and Pattern Recognition
  (CVPR)}, pages 300--309, 2023{\natexlab{b}}.

\bibitem[Liu et~al.(2024)Liu, Xu, Jin, Chen, Varma~T, Xu, and
  Su]{liu2023one2345}
Minghua Liu, Chao Xu, Haian Jin, Linghao Chen, Mukund Varma~T, Zexiang Xu, and
  Hao Su.
\newblock One-2-3-45: Any single image to 3d mesh in 45 seconds without
  per-shape optimization.
\newblock \emph{NIPS}, 36, 2024.

\bibitem[Liu et~al.(2023{\natexlab{a}})Liu, Wu, Van~Hoorick, Tokmakov,
  Zakharov, and Vondrick]{zero123}
Ruoshi Liu, Rundi Wu, Basile Van~Hoorick, Pavel Tokmakov, Sergey Zakharov, and
  Carl Vondrick.
\newblock Zero-1-to-3: Zero-shot one image to 3d object.
\newblock In \emph{International Conference on Computer Vision (ICCV)}, pages
  9298--9309, 2023{\natexlab{a}}.

\bibitem[Liu et~al.(2023{\natexlab{b}})Liu, Lin, Zeng, Long, Liu, Komura, and
  Wang]{syncdreamer}
Yuan Liu, Cheng Lin, Zijiao Zeng, Xiaoxiao Long, Lingjie Liu, Taku Komura, and
  Wenping Wang.
\newblock Syncdreamer: Generating multiview-consistent images from a
  single-view image.
\newblock \emph{arXiv preprint arXiv:2309.03453}, 2023{\natexlab{b}}.

\bibitem[Melas-Kyriazi et~al.(2023)Melas-Kyriazi, Laina, Rupprecht, and
  Vedaldi]{realfusion}
Luke Melas-Kyriazi, Iro Laina, Christian Rupprecht, and Andrea Vedaldi.
\newblock Realfusion: 360° reconstruction of any object from a single image.
\newblock In \emph{Conference on Computer Vision and Pattern Recognition
  (CVPR)}, pages 8446--8455, 2023.

\bibitem[Mildenhall et~al.(2020)Mildenhall, Srinivasan, Tancik, Barron,
  Ramamoorthi, and Ng]{mildenhall2020nerf}
Ben Mildenhall, Pratul~P. Srinivasan, Matthew Tancik, Jonathan~T. Barron, Ravi
  Ramamoorthi, and Ren Ng.
\newblock Nerf: Representing scenes as neural radiance fields for view
  synthesis.
\newblock In \emph{European Conference on Computer Vision (ECCV)}, 2020.

\bibitem[Mirzaei et~al.(2023{\natexlab{a}})Mirzaei, Aumentado-Armstrong,
  Brubaker, Kelly, Levinshtein, Derpanis, and
  Gilitschenski]{diffusion3dinpaint}
Ashkan Mirzaei, Tristan Aumentado-Armstrong, Marcus~A Brubaker, Jonathan Kelly,
  Alex Levinshtein, Konstantinos~G Derpanis, and Igor Gilitschenski.
\newblock Reference-guided controllable inpainting of neural radiance fields.
\newblock \emph{arXiv preprint arXiv:2304.09677}, 2023{\natexlab{a}}.

\bibitem[Mirzaei et~al.(2023{\natexlab{b}})Mirzaei, Aumentado-Armstrong,
  Derpanis, Kelly, Brubaker, Gilitschenski, and Levinshtein]{spinnerf}
Ashkan Mirzaei, Tristan Aumentado-Armstrong, Konstantinos~G Derpanis, Jonathan
  Kelly, Marcus~A Brubaker, Igor Gilitschenski, and Alex Levinshtein.
\newblock Spin-nerf: Multiview segmentation and perceptual inpainting with
  neural radiance fields.
\newblock In \emph{Conference on Computer Vision and Pattern Recognition
  (CVPR)}, pages 20669--20679, 2023{\natexlab{b}}.

\bibitem[Podell et~al.(2023)Podell, English, Lacey, Blattmann, Dockhorn,
  Müller, Penna, and Rombach]{sdxl}
Dustin Podell, Zion English, Kyle Lacey, Andreas Blattmann, Tim Dockhorn, Jonas
  Müller, Joe Penna, and Robin Rombach.
\newblock Sdxl: Improving latent diffusion models for high-resolution image
  synthesis.
\newblock \emph{arXiv preprint arXiv:2307.01952}, 2023.

\bibitem[Poole et~al.(2022)Poole, Jain, Barron, and
  Mildenhall]{poole2022dreamfusion}
Ben Poole, Ajay Jain, Jonathan~T. Barron, and Ben Mildenhall.
\newblock Dreamfusion: Text-to-3d using 2d diffusion.
\newblock \emph{arXiv}, 2022.

\bibitem[Qian et~al.(2024)Qian, Mai, Hamdi, Ren, Siarohin, Li, Lee,
  Skorokhodov, Wonka, Tulyakov, and Ghanem]{magic123}
Guocheng Qian, Jinjie Mai, Abdullah Hamdi, Jian Ren, Aliaksandr Siarohin, Bing
  Li, Hsin-Ying Lee, Ivan Skorokhodov, Peter Wonka, Sergey Tulyakov, and
  Bernard Ghanem.
\newblock Magic123: One image to high-quality 3d object generation using both
  2d and 3d diffusion priors.
\newblock In \emph{International Conference on Learning Representations
  (ICLR)}, 2024.

\bibitem[Radford et~al.(2021)Radford, Kim, Hallacy, Ramesh, Goh, Agarwal,
  Sastry, Askell, Mishkin, Clark, et~al.]{clip}
Alec Radford, Jong~Wook Kim, Chris Hallacy, Aditya Ramesh, Gabriel Goh,
  Sandhini Agarwal, Girish Sastry, Amanda Askell, Pamela Mishkin, Jack Clark,
  et~al.
\newblock Learning transferable visual models from natural language
  supervision.
\newblock In \emph{International Conference on Machine Learning (ICML)}, pages
  8748--8763, 2021.

\bibitem[Raj et~al.(2023)Raj, Kaza, Poole, Niemeyer, Ruiz, Mildenhall, Zada,
  Aberman, Rubinstein, Barron, et~al.]{dreambooth3d}
Amit Raj, Srinivas Kaza, Ben Poole, Michael Niemeyer, Nataniel Ruiz, Ben
  Mildenhall, Shiran Zada, Kfir Aberman, Michael Rubinstein, Jonathan Barron,
  et~al.
\newblock Dreambooth3d: Subject-driven text-to-3d generation.
\newblock \emph{arXiv preprint arXiv:2303.13508}, 2023.

\bibitem[Ravi et~al.(2020)Ravi, Reizenstein, Novotny, Gordon, Lo, Johnson, and
  Gkioxari]{ravi2020pytorch3d}
Nikhila Ravi, Jeremy Reizenstein, David Novotny, Taylor Gordon, Wan-Yen Lo,
  Justin Johnson, and Georgia Gkioxari.
\newblock Accelerating 3d deep learning with pytorch3d.
\newblock \emph{arXiv:2007.08501}, 2020.

\bibitem[Rockwell et~al.(2021)Rockwell, Fouhey, and Johnson]{pixelsynth}
Chris Rockwell, David~F. Fouhey, and Justin Johnson.
\newblock Pixelsynth: Generating a 3d-consistent experience from a single
  image.
\newblock In \emph{International Conference on Computer Vision (ICCV)}, 2021.

\bibitem[Rombach et~al.(2022)Rombach, Blattmann, Lorenz, Esser, and Ommer]{sd}
Robin Rombach, Andreas Blattmann, Dominik Lorenz, Patrick Esser, and Bj{\"o}rn
  Ommer.
\newblock High-resolution image synthesis with latent diffusion models.
\newblock In \emph{Conference on Computer Vision and Pattern Recognition
  (CVPR)}, pages 10684--10695, 2022.

\bibitem[Ruiz et~al.(2023)Ruiz, Li, Jampani, Pritch, Rubinstein, and
  Aberman]{dreambooth}
Nataniel Ruiz, Yuanzhen Li, Varun Jampani, Yael Pritch, Michael Rubinstein, and
  Kfir Aberman.
\newblock Dreambooth: Fine tuning text-to-image diffusion models for
  subject-driven generation.
\newblock In \emph{Conference on Computer Vision and Pattern Recognition
  (CVPR)}, pages 22500--22510, 2023.

\bibitem[Salimans et~al.(2016)Salimans, Goodfellow, Zaremba, Cheung, Radford,
  and Chen]{inception_score}
Tim Salimans, Ian Goodfellow, Wojciech Zaremba, Vicki Cheung, Alec Radford, and
  Xi Chen.
\newblock Improved techniques for training gans.
\newblock In \emph{Advances in Neural Information Processing Systems
  (NeurIPS)}, pages 2234--2242, 2016.

\bibitem[Sanghi et~al.(2022)Sanghi, Chu, Lambourne, Wang, Cheng, Fumero, and
  Malekshan]{clipforge}
Aditya Sanghi, Hang Chu, Joseph~G Lambourne, Ye Wang, Chin-Yi Cheng, Marco
  Fumero, and Kamal~Rahimi Malekshan.
\newblock Clip-forge: Towards zero-shot text-to-shape generation.
\newblock In \emph{Conference on Computer Vision and Pattern Recognition
  (CVPR)}, pages 18603--18613, 2022.

\bibitem[Sargent et~al.(2023)Sargent, Li, Shah, Herrmann, Yu, Zhang, Chan,
  Lagun, Fei-Fei, Sun, et~al.]{zeronvs}
Kyle Sargent, Zizhang Li, Tanmay Shah, Charles Herrmann, Hong-Xing Yu, Yunzhi
  Zhang, Eric~Ryan Chan, Dmitry Lagun, Li Fei-Fei, Deqing Sun, et~al.
\newblock Zeronvs: Zero-shot 360-degree view synthesis from a single real
  image.
\newblock \emph{arXiv preprint arXiv:2310.17994}, 2023.

\bibitem[Shi et~al.(2023{\natexlab{a}})Shi, Chen, Zhang, Liu, Xu, Wei, Chen,
  Zeng, and Su]{zero123plus}
Ruoxi Shi, Hansheng Chen, Zhuoyang Zhang, Minghua Liu, Chao Xu, Xinyue Wei,
  Linghao Chen, Chong Zeng, and Hao Su.
\newblock Zero123++: A single image to consistent multi-view diffusion base
  model.
\newblock \emph{arXiv preprint arXiv:2310.15110}, 2023{\natexlab{a}}.

\bibitem[Shi et~al.(2023{\natexlab{b}})Shi, Wang, Ye, Long, Li, and
  Yang]{mvdream}
Yichun Shi, Peng Wang, Jianglong Ye, Mai Long, Kejie Li, and Xiao Yang.
\newblock Mvdream: Multi-view diffusion for 3d generation.
\newblock \emph{arXiv preprint arXiv:2308.16512}, 2023{\natexlab{b}}.

\bibitem[Snavely et~al.(2006)Snavely, Seitz, and Szeliski]{sfm}
Noah Snavely, Steven~M Seitz, and Richard Szeliski.
\newblock Photo tourism: Exploring photo collections in 3d.
\newblock In \emph{Special Interest Group on Computer Graphics and Interactive
  Techniques (SIGGRAPH)}, pages 835--846, 2006.

\bibitem[Sohl-Dickstein et~al.(2015)Sohl-Dickstein, Weiss, Maheswaranathan, and
  Ganguli]{thermo}
Jascha Sohl-Dickstein, Eric Weiss, Niru Maheswaranathan, and Surya Ganguli.
\newblock Deep unsupervised learning using nonequilibrium thermodynamics.
\newblock In \emph{International Conference on Machine Learning (ICML)}, pages
  2256--2265, 2015.

\bibitem[Song et~al.(2020{\natexlab{a}})Song, Meng, and Ermon]{ddim}
Jiaming Song, Chenlin Meng, and Stefano Ermon.
\newblock Denoising diffusion implicit models.
\newblock \emph{arXiv preprint arXiv:2010.02502}, 2020{\natexlab{a}}.

\bibitem[Song and Ermon(2019)]{song2}
Yang Song and Stefano Ermon.
\newblock Generative modeling by estimating gradients of the data distribution.
\newblock \emph{NIPS}, 32, 2019.

\bibitem[Song et~al.(2020{\natexlab{b}})Song, Sohl-Dickstein, Kingma, Kumar,
  Ermon, and Poole]{song1}
Yang Song, Jascha Sohl-Dickstein, Diederik~P Kingma, Abhishek Kumar, Stefano
  Ermon, and Ben Poole.
\newblock Score-based generative modeling through stochastic differential
  equations.
\newblock \emph{arXiv preprint arXiv:2011.13456}, 2020{\natexlab{b}}.

\bibitem[Tancik et~al.(2023)Tancik, Weber, Ng, Li, Yi, Kerr, Wang,
  Kristoffersen, Austin, Salahi, Ahuja, McAllister, and Kanazawa]{nerfstudio}
Matthew Tancik, Ethan Weber, Evonne Ng, Ruilong Li, Brent Yi, Justin Kerr,
  Terrance Wang, Alexander Kristoffersen, Jake Austin, Kamyar Salahi, Abhik
  Ahuja, David McAllister, and Angjoo Kanazawa.
\newblock Nerfstudio: A modular framework for neural radiance field
  development.
\newblock In \emph{SIGGRAPH}, 2023.

\bibitem[Wang et~al.(2022)Wang, Du, Li, Yeh, and Shakhnarovich]{sjc}
Haochen Wang, Xiaodan Du, Jiahao Li, Raymond~A. Yeh, and Greg Shakhnarovich.
\newblock Score jacobian chaining: Lifting pretrained 2d diffusion models for
  3d generation.
\newblock \emph{arXiv preprint arXiv:2212.00774}, 2022.

\bibitem[Wang et~al.(2023)Wang, Lu, Wang, Bao, Li, Su, and
  Zhu]{wang2023prolificdreamer}
Zhengyi Wang, Cheng Lu, Yikai Wang, Fan Bao, Chongxuan Li, Hang Su, and Jun
  Zhu.
\newblock Prolificdreamer: High-fidelity and diverse text-to-3d generation with
  variational score distillation.
\newblock In \emph{Advances in Neural Information Processing Systems
  (NeurIPS)}, 2023.

\bibitem[Yang et~al.(2024{\natexlab{a}})Yang, Kang, Huang, Xu, Feng, and
  Zhao]{depthanything}
Lihe Yang, Bingyi Kang, Zilong Huang, Xiaogang Xu, Jiashi Feng, and Hengshuang
  Zhao.
\newblock Depth anything: Unleashing the power of large-scale unlabeled data.
\newblock In \emph{Conference on Computer Vision and Pattern Recognition
  (CVPR)}, 2024{\natexlab{a}}.

\bibitem[Yang et~al.(2024{\natexlab{b}})Yang, Kang, Huang, Zhao, Xu, Feng, and
  Zhao]{depth_anything_v2}
Lihe Yang, Bingyi Kang, Zilong Huang, Zhen Zhao, Xiaogang Xu, Jiashi Feng, and
  Hengshuang Zhao.
\newblock Depth anything v2.
\newblock \emph{arXiv:2406.09414}, 2024{\natexlab{b}}.

\bibitem[Yi et~al.(2024)Yi, Fang, Wang, Wu, Xie, Zhang, Liu, Tian, and
  Wang]{yi2023gaussiandreamer}
Taoran Yi, Jiemin Fang, Junjie Wang, Guanjun Wu, Lingxi Xie, Xiaopeng Zhang,
  Wenyu Liu, Qi Tian, and Xinggang Wang.
\newblock Gaussiandreamer: Fast generation from text to 3d gaussians by
  bridging 2d and 3d diffusion models.
\newblock In \emph{Conference on Computer Vision and Pattern Recognition
  (CVPR)}, 2024.

\bibitem[Zhang et~al.(2024)Zhang, Li, Wan, Wang, and Liao]{text2nerf}
Jingbo Zhang, Xiaoyu Li, Ziyu Wan, Can Wang, and Jing Liao.
\newblock Text2nerf: Text-driven 3d scene generation with neural radiance
  fields.
\newblock \emph{IEEE TVCG}, 2024.

\bibitem[Zhang et~al.(2018)Zhang, Isola, Efros, Shechtman, and Wang]{lpips}
Richard Zhang, Phillip Isola, Alexei~A Efros, Eli Shechtman, and Oliver Wang.
\newblock The unreasonable effectiveness of deep features as a perceptual
  metric.
\newblock In \emph{Conference on Computer Vision and Pattern Recognition
  (CVPR)}, pages 586--595, 2018.

\bibitem[Zhou and Tulsiani(2023)]{zhou2023sparsefusion}
Zhizhuo Zhou and Shubham Tulsiani.
\newblock Sparsefusion: Distilling view-conditioned diffusion for 3d
  reconstruction.
\newblock In \emph{Conference on Computer Vision and Pattern Recognition
  (CVPR)}, 2023.

\bibitem[Zhu et~al.(2023)Zhu, Zhuang, and Koyejo]{hifa}
Junzhe Zhu, Peiye Zhuang, and Sanmi Koyejo.
\newblock Hifa: High-fidelity text-to-3d generation with advanced diffusion
  guidance.
\newblock In \emph{International Conference on Learning Representations
  (ICLR)}, 2023.

\bibitem[Zwicker et~al.(2001)Zwicker, Pfister, Van~Baar, and Gross]{zwick}
Matthias Zwicker, Hanspeter Pfister, Jeroen Van~Baar, and Markus Gross.
\newblock Ewa volume splatting.
\newblock In \emph{Proceedings Visualization, 2001. VIS'01.,}, pages 29--538,
  2001.

\end{thebibliography}
